\documentclass{article}

\usepackage{microtype}
\usepackage{graphicx}
\usepackage{subfigure}
\usepackage{booktabs} 

\usepackage{hyperref}

\usepackage[accepted]{icml2024}

\usepackage{amsmath}
\usepackage{amssymb}
\usepackage{mathtools}
\usepackage{amsthm}
\usepackage{bm}
\usepackage{multirow}

\usepackage[capitalize,noabbrev]{cleveref}

\crefname{equation}{Eq.}{Eqs.}
\crefname{figure}{Fig.}{Figs.}
\crefname{section}{Sec.}{Secs.}

\theoremstyle{plain}
\newtheorem{theorem}{Theorem}[section]
\newtheorem{proposition}[theorem]{Proposition}

\theoremstyle{definition}

\theoremstyle{remark}

\usepackage[textsize=tiny]{todonotes}

\newcommand\Acal{\mathcal{A}}

\newcommand\p{\partial}
\newcommand\rd{{\rm d}}

\newcommand\wt{\widetilde}

\newcommand\h{\mathfrak{h}}

\icmltitlerunning{
Understanding Diffusion Models by Feynman's Path Integral
}

\begin{document}

\twocolumn[
\icmltitle{Understanding Diffusion Models by Feynman's Path Integral}




\icmlsetsymbol{equal}{*}

\begin{icmlauthorlist}
\icmlauthor{Yuji Hirono}{kyoto}
\icmlauthor{Akinori Tanaka}{aip,ithems,keio}
\icmlauthor{Kenji Fukushima}{tokyo}
\end{icmlauthorlist}

\icmlaffiliation{kyoto}{Department of Physics, Kyoto University, Kyoto 606-8502, Japan}
\icmlaffiliation{aip}{RIKEN AIP, RIKEN, Nihonbashi 103-0027, Japan}
\icmlaffiliation{ithems}{RIKEN iTHEMS, RIKEN, Wako 351-0198, Japan}
\icmlaffiliation{keio}{Department of Mathematics, Keio University, Hiyoshi 223-8522, Japan}

\icmlaffiliation{tokyo}{Department of Physics, The University of Tokyo, 7-3-1 Hongo, Bunkyo-ku, Tokyo 113-0033, Japan}


\icmlcorrespondingauthor{Yuji Hirono}{yuji.hirono@gmail.com}
\icmlcorrespondingauthor{Akinori Tanaka}{akinori.tanaka@riken.jp}
\icmlcorrespondingauthor{Kenji Fukushima}{fuku@nt.phys.s.u-tokyo.ac.jp}

\icmlkeywords{Machine Learning, ICML}

\vskip 0.3in
]



\printAffiliationsAndNotice{}

\begin{abstract}
Score-based diffusion models have proven effective in image generation and have gained widespread usage; however, the underlying factors contributing to the performance disparity between stochastic and deterministic (i.e., the probability flow ODEs) sampling schemes remain unclear. We introduce a novel formulation of diffusion models using Feynman's path integral, which is a formulation originally developed for quantum physics. We find this formulation providing comprehensive descriptions of score-based generative models, and demonstrate the derivation of backward stochastic differential equations and loss functions. The formulation accommodates an interpolating parameter connecting stochastic and deterministic sampling schemes, and we identify this parameter as a counterpart of Planck's constant in quantum physics. This analogy enables us to apply the Wentzel–Kramers–Brillouin (WKB) expansion, a well-established technique in quantum physics, for evaluating the negative log-likelihood to assess the performance disparity between stochastic and deterministic sampling schemes.
\end{abstract}

\section{Introduction}
\label{intro}

Diffusion models have demonstrated impressive performance on image generation tasks \cite{dhariwal2021diffusion} and they have earned widespread adoption across various applications \cite{yang2023diffusion}.
While the predominant contemporary application of diffusion models lies in conditional sampling driven by natural language \cite{rombach2022high}, the 
mathematical framework underlying their training, i.e.,
the score-based scheme \cite{hyvarinen2009estimation, vincent2011connection, song2019generative} or the denoising scheme \cite{sohl2015deep, ho2020denoising},
is not inherently dependent on prior conditions.
The work \cite{kingma2021variational} showed 
the equivalence of these two schemes from a variational perspective. 
The work \cite{song2020score} developed a unified description based on stochastic differential equations (SDEs).
In both cases, the sampling process is given by a Markovian stochastic process.

Another type of probabilistic models employs deterministic sampling schemes using ordinary differential equations (ODEs) such as the probability flow ODE \cite{song2020score}, which can be understood as a continuous normalizing flow (CNF) \cite{chen2018neural,lipman2022flow}.
A notable advantage of deterministic sampling schemes is that there is a bijective map between 
the latent space and the data space.
This bijection not only facilitates intricate tasks like manipulation of latent representations for image editing
\cite{su2023dual} but also enables the direct computation of negative log-likelihoods (NLLs).
Within stochastic sampling processes in contrast to deterministic ones, a direct evaluation method of the NLL remains elusive,
though a theoretical bound exists for the NLL \cite{kong2023information}.

In general, stochastic sampling schemes require more number of function evaluations (NFE) than deterministic schemes, and is inferior in terms of sample generation speed. However, the beneficial impact of stochastic generation on certain metrics, such as the Fr\'{e}chet Inception Distance (FID) \cite{heusel2017gans} is a well-known property in practice.
For example, \cite{karras2022elucidating} reports
improvements in these metrics with the incorporation of stochastic processes both in Variance-Exploding (VE) and Variance-Preserving pretrained models in \cite{song2020score}.
Intuitively, one can conceptualize noise within stochastic generation as
perturbation to propel particles out of local minima, potentially enhancing the diversity and quality of the generated samples.
However,
beyond this intuitive level,
thoroughly quantitative analysis or rigorous theoretical framework to 
explain this phenomenon is missing. 

\begin{figure}[t]
  \vskip 0.1in
  \centerline{\includegraphics[width=0.85\columnwidth]{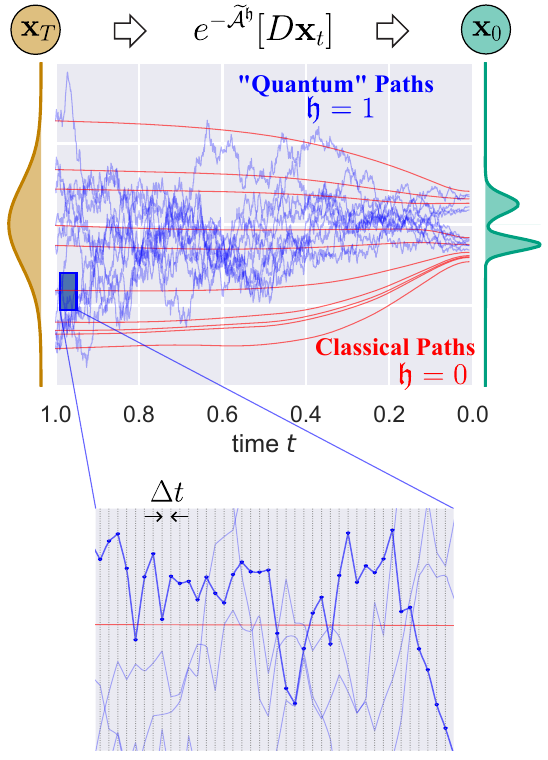}}
  \caption{ Schematic of the path integral formulation of diffusion models. }
  \label{fig:schematic}
  \vskip -0.1in
\end{figure}

We tackle these two issues by making use of the \textit{path integral formalism}, a framework originally developed in quantum physics by Feynman 
\cite{RevModPhys.20.367, Feynman:100771}.
In classical physics, a particle's motion 
draws a trajectory, i.e., a 
deterministic path in space and time. To accommodate quantum effects, the path integral formalism generalizes the trajectory including quantum fluctuations by comprising all possible paths, $\{\mathbf x_t\}_{t\in [0, T]}$, of a particle between two points, $\mathbf{x}_0$ at time $t=0$ and $\mathbf{x}_T$ at $t=T$; see a magnified panel in Fig.~\ref{fig:schematic} for a counterpart in diffusion models.
The quantum expectation value of observable $\mathcal O (\mathbf x_t)$ is computed as a weighted sum:
$\sum_{\text{paths}}
\mathcal O(\mathbf x_t) e^{i \mathcal A[\mathbf x_t] / \hbar}$, where $\hbar$ is Planck's constant.
In addition to quantum fluctuations, the path integral can be extended to incorporate stochastic fluctuations as well \cite{PhysRev.91.1505}.
We demonstrate that diffusion models can be formulated in terms of path integrals, which not only deepens our understanding about diffusion model formulations but also allows for the application of various techniques advanced in quantum physics.
Importantly, this framework provides an innovative method for calculating the NLL in stochastic generation processes of diffusion models.

Our contributions are as follows.
\begin{itemize}
\item We reformulate diffusion models using path integral techniques. We exemplify applications including 
a re-derivation of the time-reversed SDE \cite{anderson1982reverse} and an estimate
of loss functions \cite{song2021maximum} based on Girsanov's theorem \cite{oksendal2013stochastic}.
\item Following \cite{zhang2022fast}, we introduce an interpolating parameter $\h$ connecting the stochastic generation ($\mathfrak{h}=1$)
and the probability flow ODE ($\h = 0$).
In the path integral language, the limit $\h \to 0$ corresponds to the {\it classical limit} under which quantum fluctuations are dropped off.
The path integral formulation of diffusion models reveals the role of $\h$ as Planck's constant $\hbar$ in quantum physics.
\item 
We apply the Wentzel–Kramers–Brillouin (WKB) expansion \cite{messiah}, that is formulated in terms of Planck's constant $\hbar$ in quantum physics,
%
with respect to $\h$ to the likelihood calculation.
Based on the first order NLL expression, we 
quantify the merit of noise in the sampling process by 
computing the NLL as well as the 2-Wasserstein distance.
\end{itemize}
Building upon the analogy with quantum physics, these contributions
unveil a far more profound connection to physics beyond a classic viewpoint of the Brownian motion.

\section{Related works}
\label{related}

\paragraph{Diffusion models}
Key contributions in this field of diffusion models include \cite{sohl2015deep} and \cite{song2020score} which have laid the foundational principles for these models. 
Our reformulation of diffusion models by path integrals 
captures the basic mathematical characteristics of score-based diffusion models.
%
The idea of implementing stochastic variables to represent quantum fluctuations is traced back to \cite{PhysRev.150.1079}, and mathematical foundation has been established as stochastic quantization \cite{DAMGAARD1987227}.
Recent works \cite{wang2023diffusion,premkumar2023generative} suggested similarity between diffusion models and quantum physics.
However, the path integral derivation of the basic aspects of diffusion models
has not been discussed. 
There is no preceding work to explore
the WKB expansion applied in diffusion models.

\paragraph{Likelihood calculation in diffusion models}

In deterministic sampling schemes based on ODEs,
one can directly calculate log-likelihood based on the change-of-variables formula \cite{chen2018neural,song2020score,lipman2022flow}.
Once we turn on noise in the sampling process, only a formal expression is available from
the celebrated Feynman-Kac formula
\cite{huang2021variational}.
To our best knowledge, there is no stochastic case where the values of log-likelihood have ever been calculated.
Our approach of the perturbative expansion with respect to 
$\h$ can access the log-likelihood explicitly even in the presence of noise. 
Note that an exact formula for the log-likelihood 
has been derived based on the information theory connecting to the Minimum Mean Square Error (MMSE) regression \cite{kong2023information}.
However, this method still necessitates the computation of expectation values and cannot be implemented using a single trajectory.

\paragraph{Stochastic and deterministic sampling procedures}

The role of stochasticity in the sampling process has been investigated in  
\cite{karras2022elucidating}
through empirical studies.
The gap between the score-matching objective 
depending on the sampling schemes has been pointed out in \cite{NEURIPS2021_c11abfd2}. 
To bridge the gap between the model distribution generated by the probability flow and the actual data distribution, 
\cite{pmlr-v162-lu22f}
introduced a higher-order score-matching objective. 
Besides these efforts, 
\cite{pmlr-v202-lai23d} derived an equation to be satisfied by a score function and introduced a regulator in the loss function to enforce this relation. 
The present approach is complementary to these works; we employ the perturbative $\h$-expansion and directly evaluate the noise-strength dependence of the NLL for pretrained models.

\section{Reformulation of diffusion models by path integral formalism}
\label{path-int}

In this section, we describe the reformulation of score-based diffusion models in terms of path integrals.

\subsection{ Forward and reverse processes }

For a given datapoint $\mathbf{x}_0$ sampled from an underlying data distribution $p_0$,
the first step of a score-based diffusion model is to gradually modify the data by adding noise 
via a forward SDE,
\begin{align}
  \rd \mathbf{x}_t = \bm{f}(\mathbf{x}_t, t) \rd t + g(t) \rd\mathbf{w}_t
  .
  \label{eq:sde}
\end{align}
One can view this diffusion process as a collection of stochastic trajectories, 
and consider the ``path-probability'' $P(\{\mathbf{x}_t\}_{t \in [0, T]})$ associated with Eq.~\eqref{eq:sde}.
Intuitively, it corresponds to the joint probability for the ``path'' $\{\mathbf{x}_t\}_{t \in [0, T]}$
(see also Fig.~\ref{fig:schematic} for schematic illustration of paths).
In the path integral formulation of quantum mechanics, the expectation value of observables is expressed as a summation over all possible paths weighted by an exponential factor with a quantity
called an {\it action}.
%
We observe that
the process \eqref{eq:sde} of diffusion models can be represented as a path integral:
\begin{proposition}
  \label{thm:path-integral}
The path-probability $P(\{\mathbf{x}_t\}_{t \in [0, T]})$ can be represented in the following path integral form:
\begin{align}
 P(\{\mathbf{x}_t\}_{t \in [0, T]})
 &=
 p_0(\mathbf{x}_0) e^{- \Acal } [D\mathbf{x}_t] ,
 \label{eq:path-prob}
\end{align}
with $\mathcal A \coloneqq 
\int_0^T L(\dot{\mathbf{x}}_{t}, \mathbf{x}_{t}) \rd t + J$, 
where $L(\dot{\mathbf{x}}_t, \mathbf{x}_t)$ is called Onsager–Machlup function \cite{PhysRev.91.1505} defined by
\begin{align*}
 L(\dot{\mathbf{x}}_t, \mathbf{x}_t)
 \coloneqq
 \frac{ \| \dot{\mathbf{x}}_t - \bm{f}(\mathbf{x}_t, t)\|^2}{2g(t)^2 } ,
\end{align*}
and $J$ is the Jacobian
associated with the chosen discretization scheme in stochastic process.
\end{proposition}
Here, the overdot indicates the time derivative.
In the physics literature, 
$L(\dot{\mathbf{x}}_t, \mathbf{x}_t)$ is called the \textit{Lagrangian} and $\mathcal A$ is the \textit{action}.
For a detailed derivation of \cref{eq:path-prob} and the explicit expression for $J$, see \cref{app:deriv-path-int}.

Using the path probability \eqref{eq:path-prob}, 
the expectation value of 
any observable $\mathcal O(\mathbf x_t)$
depending on $\mathbf x_t$ obeying \cref{eq:sde} is represented as
$
\mathbb E [\mathcal \mathcal O(\mathbf x_t)]
= 
\int O(\mathbf x_t)
P(\{\mathbf{x}_t\}_{t \in [0, T]})
$.
This expression is commonly referred to as a {\it path integral}
as it involves the summation over infinitely many paths.

Here, we present an intuitive explanation of the expression~\eqref{eq:path-prob}.
Let us start from a discretized version of the SDE \eqref{eq:sde} by Euler-Maruyama scheme:
\begin{align*}
  &\mathbf{x}_{t+\Delta t} 
  = \mathbf{x}_t + \bm{f}(\mathbf{x}_t, t) \Delta t + g(t)\sqrt{\Delta t} \mathbf{v}_t,
  \quad
  \mathbf{v}_t \sim \mathcal{N}(\mathbf{0}, I),
  \\
  &\Leftrightarrow
  p
  (\mathbf{x}_{t+\Delta t}|\mathbf{x}_t)
  = \mathcal{N}(\mathbf{x}_{t+\Delta t}|\mathbf{x}_t + \bm{f}(\mathbf{x}_t, t) \Delta t, g(t)^2 \Delta t I)
  ,
\end{align*}
where $\Delta t$ is time interval and $I$ is the identity matrix.
Now the time evolution of the SDE is described by a conditional Gaussian distribution. In the following, we take $\mathbf{x}_{t+\Delta t} - \mathbf{x}_t \approx \dot{\mathbf{x}}_t \Delta t$, and deform the conditional Gaussian as
\begin{align*}
  p
  (\mathbf{x}_{t+\Delta t}|\mathbf{x}_t)
  &\propto
  e^{
    - 
      \frac{\| \dot{\mathbf{x}}_t - \bm{f}(\mathbf{x}_t, t) \|^2}{2g(t)^2 }
    \Delta t 
  }
  =
  e^{
    - L(\dot{\mathbf{x}}_t, \mathbf{x}_t)
    \Delta t 
  }
  .
\end{align*}
Now let us consider the probability for realizing an explicit ``path'' $\{\mathbf{x}_t\} := [\mathbf{x}_0, \mathbf{x}_{\Delta t},  \mathbf{x}_{2 \Delta t}, \cdots,  \mathbf{x}_{T}]$, and regarding the summation as the Riemannian sum, we get the following path integral expression for the path probability:
\begin{equation}
\begin{split}
  &
  p_0(\mathbf{x}_0) 
  p
  (\mathbf{x}_{\Delta t}|\mathbf{x}_0)
  \cdots
  p
  (\mathbf{x}_{T}|\mathbf{x}_{T-\Delta t})
  \prod_{t} \rd \mathbf{x}_t
  \\
  &\eqqcolon  
  p_0(\mathbf{x}_0) 
  e^{
    - \sum_{n=0}^{T/\Delta t-1}
    L(\dot{\mathbf{x}}_{n\Delta t}, \mathbf{x}_{n\Delta t})
    \Delta t
  }
  [D\mathbf{x}_t]_{\Delta t}
  ,
  \label{eq:disc-path-prob}
\end{split}
\end{equation}
where 
$[D\mathbf{x}_t]_{\Delta t}$ 
contains a normalization constant depending on the discretization step $\Delta t$. 
We take the $\Delta t\to 0$ limit at the end of the calculation and omit $\Delta t$ in the subscript in later discussions.
In the expression~\eqref{eq:disc-path-prob},
we need to include an additional contribution 
denoted by $J$ in \cref{thm:path-integral}, depending on the choice of the discretization scheme (see \cref{app:deriv-path-int} for details).

The sampling process of score-based diffusion models
is realized by the time-reversed version 
of the forward SDE~\eqref{eq:sde}.
The path integral reformulation is beneficial in furnishing an alternative derivation of the reverse-time SDE:
\begin{proposition}
\label{thm:reverse_sde}
Let $P(\{\mathbf{x}_t\}_{t \in [0, T]})$ be the path-probability, and $p_t(\mathbf{x})$ be the distribution at time $t$ determined by the Fokker-Planck equation corresponding to \cref{eq:sde}.
The path-probability is written 
using the reverse-time action $\wt \Acal$ as
\begin{align*}
  P(\{\mathbf{x}_t\}_{t \in [0, T]})
  &= 
  e^{
  - \wt \Acal
  }\, p_T(\mathbf{x}_T) [D\mathbf{x}_t],
\end{align*}
with 
$\wt\Acal \coloneqq 
\int_0^T \widetilde{L}(\dot{\mathbf{x}}_{t}, \mathbf{x}_{t}) \rd t + \widetilde{J}
$, where 
\begin{align*}
  \widetilde{L}(\dot{\mathbf{x}}_{t}, \mathbf{x}_{t})
  &\coloneqq
  \frac{ \| \dot{\mathbf{x}}_t - \bm{f}(\mathbf{x}_t, t) + g(t)^2 \nabla \log p_t(\mathbf{x}_t) \|^2}{2g(t)^2 }
  ,
\end{align*}
and $\widetilde{J}$ is the Jacobian for reverse process depending on the discretization scheme.
\end{proposition}
We provide the proof in \cref{app:derivation-reverse-sde}.
We emphasize that the path integral derivation does not rely on the reverse-time SDE \cite{anderson1982reverse}. 
In fact, $\widetilde{L}$ involves the score-function $\nabla \log p_t(\mathbf{x}_t)$, so that \cref{thm:reverse_sde} gives us another derivation of the reverse-time SDE,
\begin{align}
  \rd \mathbf{x}_t = [\bm{f}(\mathbf{x}_t, t) - g(t)^2 \nabla \log p_t(\mathbf{x}_t)] \rd t + g(t) \rd\bar{\mathbf{w}}_t, 
  \label{eq:resde}
\end{align}
by inverting the discussion from \cref{eq:sde} to \cref{thm:path-integral}.

\subsection{Models and objectives}

In score-based models \cite{song2019generative,song2021maximum, song2020score},  
the score function $\nabla \log p_t (\mathbf x_t)$ is approximated by 
a neural network $\bm s_\theta (\mathbf x_t, t)$, 
and samplings are performed based on 
\begin{align}
  \rd \mathbf{x}_t = 
  [\bm{f}(\mathbf{x}_t, t) - g(t)^2 \bm{s}_\theta(\mathbf{x}_t, t) ] \rd t + g(t) \rd\bar{\mathbf{w}}_t
  ,
  \label{eq:rescoresde}
\end{align}
that is a surrogate for the reverse-time~\eqref{eq:resde}.
By repeating the same argument as \cref{thm:reverse_sde}, 
we can conclude that the path-probability $Q_\theta(\{\mathbf{x}_t\})$  corresponding
to \cref{eq:rescoresde} takes the following path integral representation:
\begin{align}
  Q_\theta(\{\mathbf{x}_t\})
  &=
  e^{-\wt \Acal_\theta
  } \pi(\mathbf{x}_T) [D\mathbf{x}_t],
\end{align}
where $\pi(\cdot)$ is a prior distribution, typically chosen to be the standard normal distribution.  Here,
$\wt \Acal_\theta
\coloneqq 
\int_0^T \widetilde{L}_\theta(\dot{\mathbf{x}}_{t}, \mathbf{x}_{t}) \rd t + \widetilde{J}_\theta
$
with $\widetilde{L}_\theta$ the Onsager–Machlup function for the SDE~\eqref{eq:rescoresde} defined by
\begin{align}
  \widetilde{L}_\theta(\dot{\mathbf{x}}_{t}, \mathbf{x}_{t})
  &\coloneqq
  \frac{\| \dot{\mathbf{x}}_t - \bm{f}(\mathbf{x}_t, t) + g(t)^2 \bm{s}_\theta(\mathbf{x}_t, t) \|^2}{2g(t)^2 }
  ,
\end{align}
and $\wt{J}_\theta$ is the Jacobian contribution.
Figure~\ref{fig:schematic} depicts a schematic picture of these reverse-time processes.

Now, we can follow the training scheme based on bound of the log-likelihood or equivalently, KL divergence via data-processing inequality \cite{song2021maximum}:
\begin{align}
  D_{\rm KL}(p_0\|{q_\theta}_0)
  \leq
  D_{\rm KL}(P\|Q_\theta)
  .
  \label{eq:data-processing}
\end{align}
The r.h.s.\ of \cref{eq:data-processing} can be calculated by using \textit{Girsanov's theorem} \cite{oksendal2013stochastic}. 
An equivalent computation can also be performed in the path integral formulation:
\begin{proposition}
  \label{thm:KL}
  The KL divergence of path-probabilities can be represented by the path integral form,
  \begin{align*}
    D_{\rm KL}(P\|Q_\theta)
    &=
    \int 
    e^{- \wt \Acal
    }  
    p_T(\mathbf{x}_T)
    \log \frac{
      e^{- \wt \Acal }\,
      p_T(\mathbf{x}_T)
    }{
      e^{- \wt{\Acal}_\theta }
      \pi(\mathbf{x}_T)
    }[D\mathbf{x}_t],
  \end{align*}
  and it can be computed as
  \begin{align*}
    &D_{\rm KL}(P\|Q_\theta)
    =
    D_{\rm KL}(p_T \| \pi)
    \\
    &\quad \quad 
    +
    \int_0^T \frac{g(t)^2}{2}
    \mathbb{E}_{p_t}
    \| \nabla \log p_t(\mathbf{x}_t) - \bm{s}_\theta(\mathbf{x}_t, t)\|^2
    \rd t. 
  \end{align*}
\end{proposition}
Indeed, \cref{thm:KL} yields the same contribution as the calculation based on Girsanov's theorem.
We give the derivation of \cref{thm:KL} in \cref{app:eval-kl}.

The discussion so far can be straightforwardly extended to the cases with fixed initial conditions. 
One can basically make the replacement,
$p_t (\mathbf x_t) \to p_t (\mathbf x_t | \mathbf x_0)$; see \cref{app:conditional} for details.

\section{Log-likelihood by WKB expansion}

We have so far discussed reformulation 
of score-based diffusion models 
using the path integral formalism. 
This reformulation allows for the techniques developed in quantum physics for analyzing the properties of diffusion models. 
As an illustrative example, we present the calculation of log-likelihood in the presence of noise in the sampling process by pretrained models.

\subsection{Interpolating SDE and probability flow ODE}

Following \cite{zhang2022fast}, we consider a family of generating processes defined by the following SDE parametrized by $\h \in \mathbb R_{\ge 0}$:
\begin{align}
  \rd \mathbf x_t = \left[
  \bm{f}(\mathbf x_t, t) - \frac{1+{\h}}{2}g(t)^2 \bm{s}_\theta(\mathbf{x}_t, t)
  \right] \rd t + {\sqrt{\h}} g(t) \rd \bar{\mathbf{w}}_t
  .
  \label{eq:1para_gen_sde}
\end{align}
If we take $\h = 0$, the noise term vanishes, and the process reduces to the \textit{probability flow ODE} \cite{song2020score}.
The situation at $\h=1$  corresponds to the original SDE \eqref{eq:rescoresde}.
%
%
The path-probability corresponding to this process~\eqref{eq:1para_gen_sde} can be expressed as 
\begin{align}\label{eq:path-prob-h}
Q_\theta^\h(\{\mathbf{x}_t\}_{t \in [0, T]})
&= 
e^{-\wt{\Acal}_\theta^\h 
} \pi(\mathbf{x}_T) [D\mathbf{x}_t],
\end{align}
with 
$
\wt{\Acal}_\theta^\h 
\coloneqq 
\int_0^T \widetilde{L}_\theta^\h(\dot{\mathbf{x}}_{t}, \mathbf{x}_{t}) \rd t + \widetilde{J}_\theta^\h
$,
where 
\begin{equation}
\widetilde{L}_\theta^\h(\dot{\mathbf{x}}_{t}, \mathbf{x}_{t})
\coloneqq
\frac{\| \dot{\mathbf{x}}_t - \bm{f}(\mathbf{x}_t, t) + \frac{1+\h}{2}g(t)^2 \bm{s}_\theta(\mathbf{x}_t, t)\|^2}{2 \h g(t)^2}
,
\label{eq:o-m-func-q-h}
\end{equation}
and $\widetilde{J}_\theta^\h$ is the Jacobian contribution.

The way how $\h$ enters \cref{eq:o-m-func-q-h} is quite suggestive: the action 
$\wt{\Acal}^\h_\theta$ is inversely proportional to $\h$, similarly to quantum mechanics where
the path integral weight takes a form of $e^{- i \Acal / \hbar }$.
This structural similarity provides us with a physical interpretation of the limit, $\h \to 0$; it realizes the {\it classical limit} in the path integral representation. 
Moreover, this analogy between $\h$ and $\hbar$ naturally leads us to explore a perturbative expansion in terms of $\h$, i.e., the {\it WKB expansion}, a well-established technique to treat the asymptotic series expansion in mathematical physics. 

\begin{figure}[t]
  \vskip 0.1in
  \centerline{\includegraphics[width=0.85\columnwidth]{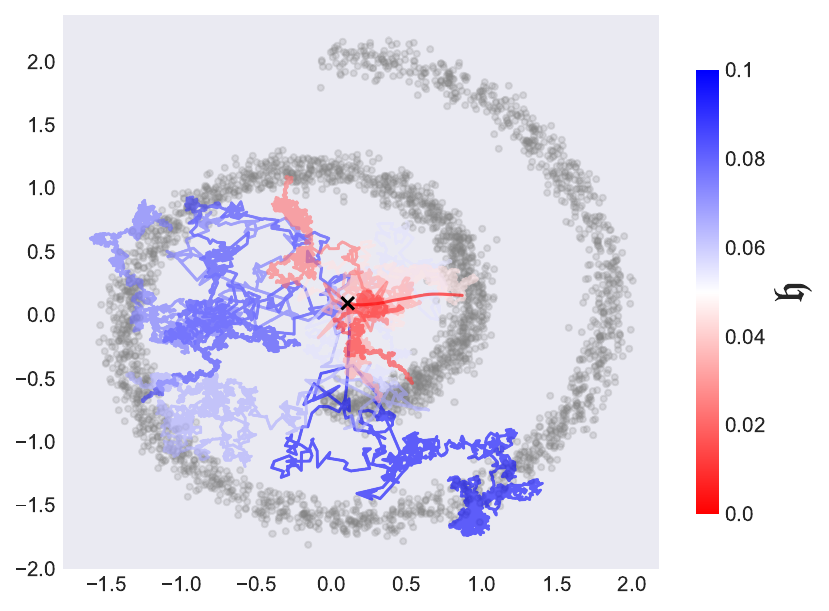}}
  \caption{\textbf{Gray dots:} training data. \textbf{Colored lines:} generative trajectories based on \cref{eq:1para_gen_sde} for different noise levels $\h$ from the identical initial vector $\mathbf{x}_T$ shown by $\times$.}
  \label{fig:interpolation}
  \vskip -0.1in
\end{figure}

In the classical limit of $\h \to 0$, the dominant contribution comes from the path that minimizes the action, realizing the \textit{principle of least action} in the physics context.
Since the Lagrangian $\wt{L}^\h_\theta$ of this model is nonnegative,
$e^{-\wt\Acal^\h_\theta}$ goes vanishing in the limit $\h \to 0$ unless the path satisfies
$\wt{L}^{\h=0}_\theta = 0$ or equivalently 
$\dot{\mathbf x} = \bm{f}_\theta^\text{PF}(\mathbf{x}_t, t)$, where
$\bm{f}_\theta^\text{PF}(\mathbf{x}_t, t)$ is the drift term for the probability flow ODE defined by
\begin{align}
  \bm{f}_\theta^\text{PF}(\mathbf{x}_t, t)
  \coloneqq 
  \bm{f}(\mathbf{x}_t, t) - \frac{1}{2}g(t)^2 \bm{s}_\theta(\mathbf{x}_t, t)
  .
  \label{eq:f_pf}
\end{align}
Consequently, the path-probability reduces to the product of delta functions as $\h \to 0$:
\begin{align*}
  &e^{- \wt{\Acal}_\theta^\h 
  } \pi(\mathbf{x}_T) [D\mathbf{x}_t]
  \\
  &\to
  \left(
  \prod_{t=0}^T\delta\bigl(\dot{\mathbf{x}}_t - \bm{f}_\theta^\text{PF}(\mathbf{x}_t, t)\bigr) 
  \right) 
  e^{-\tilde{J}^{\h=0}_\theta}
  \pi(\mathbf{x}_T) [D\mathbf{x}_t].
\end{align*}
This situation is visualized in \cref{fig:interpolation}:
we plot trajectories 
generated by the SDE \eqref{eq:1para_gen_sde} from fixed $\mathbf{x}_T$ with various $\h$s 
with a pretrained model.
The trajectories are concentrated near the ODE path 
when the noise level is low ($\h \approx 0$), which means that the path-probability for $\h \to 0$ reduces the classical path represented by Dirac's delta function. 
In this way, 
the realization of the probability flow ODE in $\h \to 0$ can be regarded as a reminiscent of the reduction from quantum mechanics to classical mechanics in $\hbar \to 0$.
The 1-dim.\ trajectories for $\h=0$ (classical paths) and $\h=1$ (``quantum'' paths) are also visualized in \cref{fig:schematic}.

For $\h=0$, it is well-known that the log-likelihood can be written exactly.
Employing the It\^o scheme, we can recover the log-likelihood by the path integral with fixed initial condition with $\mathbf{x}_0$ as 
\begin{align}
  &\log q_\theta^{\h=0}(\mathbf{x}_0)
  \notag
  \\
  &=
  \log \int_{\mathbf{x}_0} 
  \prod_{t=0}^T\delta\bigl(\dot{\mathbf{x}}_t - \bm{f}_\theta^\text{PF}(\mathbf{x}_t, t)\bigr) 
  e^{-\tilde{J}^{\h=0}_\theta}
  \pi(\mathbf{x}_T) [D\mathbf{x}_t]
  \notag
  \\
  &=
  \log 
  \pi(\mathbf{x}_T)
+ \int_0^T \nabla \cdot \bm{f}_\theta^\text{PF}(\mathbf{x}_t, t) \rd t
  ,
  \label{eq:log-likelihood_pf}
\end{align}
where $\mathbf{x}_T$ in the last line 
is obtained from the solution of $\dot{\mathbf{x}}_t = \bm{f}_\theta^\text{PF}(\mathbf{x}_t, t)$ at time $T$, 
and we have used the explicit expression of ${\tilde{J}^{\h=0}_\theta}$
(see \cref{app:to_path-integral}).
Equation \eqref{eq:log-likelihood_pf} is nothing but the instantaneous change-of-variables formula of an ODE flow~\cite{chen2018neural}.
The computations presented here can be equivalently performed in different discretization schemes, and it should be noted that the final results are free from such scheme dependences.

\subsection{$\h \neq 0$ in path integral}

When the score estimation is imperfect,
the probability distribution $q^\h_0 (\mathbf x_0)$ of a model 
acquires nontrivial dependency on parameter $\h$. 
This implies that the quality of sampled images varies depending on the level of noisiness parametrized by $\h$.

As we discussed earlier, in quantum physics,
$\hbar \to 0$ corresponds to the classical limit, 
and the effect of small but nonzero $\hbar$ 
can be taken into account
as a series expansion with respect to $\hbar$, 
commonly referred to as the WKB expansion or the semi-classical approximation.
In the path integral formulation of diffusion models, $\h$ plays a role of 
Planck's constant $\hbar$, 
which quantifies the degrees of ``quantumness.''
As a basis for the WKB expansion
of the log-likelihood, we have found the following result: 
\begin{theorem}\label{thm:likelihood_wkb}
The log-likelihood for the process~\eqref{eq:1para_gen_sde} satisfies
\footnotesize
\begin{equation}
\begin{split}
&\log q^\h_0 (\mathbf x_0) 
= 
\log \pi  (\mathbf x_T) 
\\
&
+ 
\int_0^T \rd t 
\, 
\nabla \!\cdot\!
\Bigl(
\bm f^{\rm PF}_{\theta} (\mathbf x_t, t)
- 
\frac{\h g(t)^2}{2}
[
\bm s_{\theta} (\mathbf x_t, t) - \nabla \log q^\h_t(\mathbf x_t)
]
\Bigr) ,
\end{split}
\label{eq:q-h}
\end{equation}
\normalsize
where $\pi(\cdot)$ is a prior and 
$\mathbf{x}_t$ is the solution for the modified probability flow ODE,
\begin{equation}
\dot{\mathbf{x}}_t 
=
\bm f^{\rm PF}_{\theta} (\mathbf x_t,t)
- 
\frac{\h g(t)^2}{2} 
[
\bm s_{\theta}(\mathbf x_t, t)
- 
\nabla \log q^\h_t (\mathbf x_t) 
],
\label{eq:modified-p-flow}
\end{equation}
with initial condition $\mathbf x_{t=0} = \mathbf x_0$. 
\end{theorem}
We provide the proof of this theorem in \cref{app:proof-q-h-formula}

Note that \cref{eq:q-h} is valid for any value of $\h\in\mathbb{R}_{\ge 0}$.
The modified probability flow ODE \eqref{eq:modified-p-flow} indicates that, 
when the score deviates from the ground-truth value, the deterministic trajectory should also be modified by $\h \neq 0$.
In \cref{eq:q-h}, the density $q^\h_t (\mathbf x)$ appears on both sides and this is a self-consistent equation.
This relation provides us with the basis for perturbative evaluation of the NLL in power series of $\h$. 

\begin{theorem}\label{thm:likelihood_perturbative}
The perturbative expansion of the log-likelihood \eqref{eq:1para_gen_sde} 
to the first order in $\h$ reads
\footnotesize
\begin{equation}
  \begin{split}
  &\log q^\h_0 (\mathbf x_0) 
  = 
  \log q^{\h=0}_0 (\mathbf x_0) 
  \\
  &
  + \h \Bigg(
    \delta \mathbf x_T \cdot \nabla \log \pi (\mathbf x_T)
    + \int_0^T \rd t 
        \, 
        \nabla \cdot \delta {\bm f}^{\rm PF}_{\theta}(\mathbf x_t, \delta \mathbf x_t, t)
  \Bigg)
  ,
  \end{split}
  \label{eq:q-h-1}
  \end{equation}
  \normalsize
  where $(\mathbf{x}_t, \delta \mathbf{x}_t)$ is the solution for the coupled probability flow ODE,
  \begin{equation}
    \begin{split}
  \dot{\mathbf{x}}_t 
  &=
  \bm f^{\rm PF}_{\theta} (\mathbf x_t,t),
  \\
  \dot{\delta \mathbf x}_t 
  &=
  \delta {\bm f}^{\rm PF}_{\theta}(\mathbf x_t, \delta \mathbf x_t, t)
  ,
  \end{split}
  \label{eq:coupled-p-flow}
  \end{equation}
  with initial condition $\mathbf x_{t=0}=\mathbf x_0$ and $\delta \mathbf x_{t=0} = \mathbf 0$, 
  where
  \footnotesize
  \begin{align*}
    &\delta {\bm f}^{\rm PF}_{\theta}(\mathbf x_t, \delta \mathbf x_t, t)
    \notag\\
    &\coloneqq
    (\delta \mathbf x_t \cdot \nabla) \bm f^{\rm PF}_{\theta} (\mathbf x_t, t) - \frac{g(t)^2}{2} [
      \bm s_{\theta}(\mathbf x_t, t)
      - 
      \nabla \log q^{\h=0}_t (\mathbf x_t) 
    ]
  .
  \label{eq:delta-p-flow}
  \end{align*}
  \normalsize
\end{theorem}
We provide the proof in \cref{app:likelihood_perturbative}.
Note that we treat $\mathbf x_t$ and $\delta \mathbf x_t$ as \textit{independent} variables, so the gradient $\nabla$ does not act on $\delta \mathbf x_t$.
Because \cref{eq:q-h-1} is no longer a self-consistent equation,
this theorem allows us to compute $O(\h^1)$ correction of the log-likelihood based on $\log q^{\h=0}_t(\mathbf x_t)$, i.e., log-likelihood defined by the usual probability flow ODE.

A unique feature of the $\h$-expansion in this model 
is that $\h$ appears in the denominator of 
\cref{eq:o-m-func-q-h} as well as in the numerator, 
which contrasts quantum physics 
where $\hbar$ only appears in the denominator as an overall factor.
As a result, once we consider finite $\h$ corrections, 
the ``classical'' path deviates from the classical path obtained in the $\h\to 0$ limit. 
This deviation caused by $\h\neq 0$ is quantified by $\delta \mathbf x_t$, which
represents how noise influences the bijective relationship between the data points and their latent counterparts.

\section{Experiments}
\label{exp}

\subsection{ Analysis of 1-dim.\ Gaussian data }\label{sec:1d-gauss}

Let us first illustrate the results in a simple example of one-dimensional (1-dim.) Gaussian distribution, in which everything is analytically tractable.
We take the data distribution to be 1-dim.\ Gaussian distribution with zero mean and variance $v_0$. 
For the forward process, 
we employ $f = - \beta x / 2$ and $g(t) = \sqrt{\beta}$, which corresponds to {\it denoising diffusion probabilistic model} (DDPM) for general $\beta$ \cite{ho2020denoising}.
Here, we specifically take $\beta$ to be constant for simplicity. 
To study the relation between the imperfect score estimation and stochasticity in the sampling process, we parametrize the score model as
$s(x,t)=- x (1 + \epsilon)/v_t$,
where $\epsilon \in \mathbb R$ quantifies the deviation from the perfect score,
and $v_t$ is the variance of the distribution in the forward process. 
In this simple model, the distribution remains Gaussian in both forward and backward processes.
When $\epsilon = 0$, the backward time evolution exactly matches the forward one, and 
$q_t (x) = \mathcal N (x | 0, v_t)$. 
When $\epsilon \neq 0$, the model distribution $q^\h_t (x)$ nontrivially depends on $\h$. 
We evaluate $q^\h_t (x)$ and examine the effect of noisiness in the sampling process. 
In this model, we can compute $q^\h_t (x)$ analytically and verify that \cref{eq:q-h} is satisfied, as we detail in \cref{app:ex-gaussian}.

\begin{figure}
    \centerline{\includegraphics[width=0.98\columnwidth]{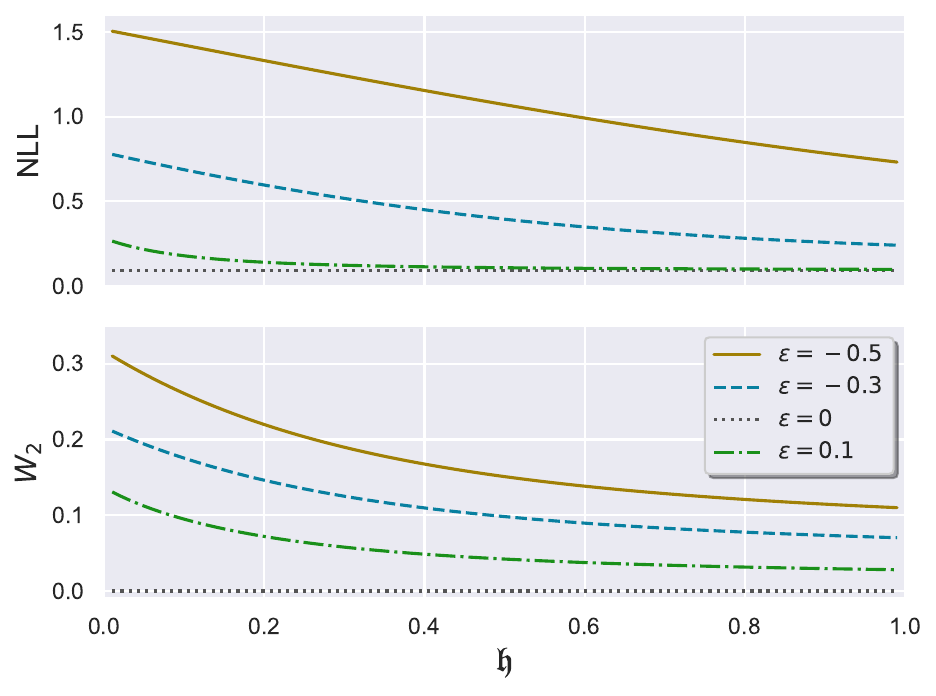}}
    \caption{\textbf{(Top)} Negative log-likelihood (NLL) of the 1-dim.\ Gaussian toy model. \textbf{(Bottom)} 2-Wasserstein metric ($W_2$) between the data distribution and the distribution obtained by the same model.  Both panels are plotted as a function of parameter $\mathfrak{h}$.}
    \label{fig:ce_and_w2-gauss}
\end{figure}

In \cref{fig:ce_and_w2-gauss} {(Top)}, we plot the NLL, $- \mathbb E_{p_0} [\log q^\h_0(x_0)]$, as a function of parameter $\h$. Different lines correspond to different values of $\epsilon$. 
The NLL is computed analytically, and we give the details in \cref{app:ex-gaussian}.
For nonzero $\epsilon$, the score estimation is imperfect and the model distribution $q^\h_0 (x_0)$ differs from the data distribution. 
In these situations at $\epsilon\neq 0$, the NLL acquires nontrivial dependence on $\h$, meaning that the quality of generated images depends the noise level in the sampling process. 
In \cref{fig:ce_and_w2-gauss} {(Top)}, the NLLs are decreasing functions of $\h$, and the presence of noise improves the output quality. 
Depending on the choice of $\epsilon$, 
the NLL could be an increasing function of $\h$ as well. 
For reference,
in \cref{fig:ce_and_w2-gauss} (Bottom),
we plot the 2-Wasserstein distance, $W_2$, between the data distribution and $q^\h_0(x_0)$ for different values of $\epsilon$. 
The qualitative behavior is consistent with that of the NLL.

\subsection{Experiment of 2-dim. synthetic data}
\label{sec:2d-synthetic}
Let us show some examples of log-likelihood computation for noisy generating process with pre-trained diffusion models trained by two-dimensional (2-dim.) synthetic distributions, Swiss-roll and 25-Gaussian (see \cref{app:data}).

\paragraph{Pretrained models}
We train simple neural network: $\mathbf (\mathbf x, t) \to {\tt concat}([\mathbf x, t]) \to \big({\tt Dense}(128) \to {\tt swish} \big)^3 \to {\tt Dense}(2) \to {\bm s}_\theta(\mathbf x, t)$ by choosing one of the following SDE schedulings:
\begin{description}
  \item[\begin{sc}simple\end{sc}:] 
  $ \scriptsize
  \bm f(\mathbf x, t) = -\frac{\beta}{2}t \mathbf x, 
  \quad
  g(t)= \sqrt{\beta t},
  \quad
  (\beta = 20)
  $
  \item[\begin{sc}cosine\end{sc}:] 
  $
  \scriptsize
  \bm f(\mathbf x, t) = -\frac{\pi}{2} \tan (\frac{\pi}{2}t) \mathbf x, 
  \quad
  g(t)= \sqrt{\pi \tan (\frac{\pi}{2}t)},
  \normalsize
  $
\end{description}
with time interval $t \in [T_\text{min}, T_\text{max}]$, where $T_\text{min} \in \mathbb{R}$ is small but nonzero to avoid singular behavior around $t=0$.
We take $T_\text{min}=0.01,\, T_\text{max}=1$ here.
We train each model by using the following loss function:
\begin{equation}
    \frac{1}{N_\text{batch}} \sum_{i=1}^{N_\text{batch}}
  \frac{g(t_i)^2}{2} 
  \left\|
  \frac{\mathbf x_{i, t} - \alpha(t_i)\mathbf x_i}{\sigma(t_i)^2}
  + {\bm s}_\theta(\mathbf x_{i, t}, t_i)
  \right\|^2 
  ,
  \end{equation}
where the signal and noise functions $\alpha(t)$ and $\sigma(t)$ are fixed by the chosen SDE scheduling, and $t_i$ represents discretized time.
To compute the above, we take the Monte-Carlo sampling with $N_{\text{batch}}$ batches of data according to
$\mathbf x_{i, t} \sim \mathcal{N}(\alpha(t_i)\mathbf x_i, \sigma(t_i)^2 I)$.
For more details about pretraining with explicit forms of $\alpha(t)$, $\sigma(t)$, $t_i$, and $N_{\text{batch}}$, see explanations around \cref{eq:loss_MC} in \cref{app:training}.
%

\paragraph{NLL calculation}
Our basic strategy is based on \cref{thm:likelihood_perturbative}, the perturbative expansion of the log-likelihood with respect to the parameter $\h$.
To calculate the $O(\h)$ correction, we need the values of $\log q^{\h=0}_t(\mathbf x)$ and its derivatives.
We can obtain $\log q^{\h=0}_t(\mathbf x)$ by solving the probability flow ODE; however, there is no closed formula for its derivatives. 
In this experiment, the dimension of $\mathbf x = [x, y]^\top$ is 2,
and this relatively low dimensionality allows us 
to approximate derivatives by discretized differential operators with small $\Delta x$:
\begin{align}
  &\hat{\nabla} \log q^{\h=0}_t(\mathbf x)
  \notag \\
  &=
  \left[
  \begin{array}{l}
    \frac{
      \log q^{\h=0}_t(\mathbf x_t + [\Delta x, 0]^\top)
      -
      \log q^{\h=0}_t(\mathbf x_t - [\Delta x, 0]^\top)
      }{2\Delta x}
    \\
    \frac{
      \log q^{\h=0}_t(\mathbf x_t + [0, \Delta x]^\top)
      -
      \log q^{\h=0}_t(\mathbf x_t - [0, \Delta x]^\top)
      }{2\Delta x}
  \end{array}
  \right],
  \label{eq:discrete_nabla}
  \\
  &\hat{\nabla}^2 \log q^{\h=0}_t(\mathbf x)
  \notag \\
  &=
  \Big(
    \log q^{\h=0}_t(\mathbf x_t + [\Delta x, 0]^\top )
     + \log q^{\h=0}_t(\mathbf x_t - [\Delta x, 0]^\top)
     \notag\\&\quad
     +\log q^{\h=0}_t(\mathbf x_t + [0, \Delta x]^\top)
     +\log q^{\h=0}_t(\mathbf x_t - [0, \Delta x]^\top)
     \notag\\&\quad
     -4\log q^{\h=0}_t(\mathbf x_t)
    \Big)/\Delta x^2 .
  \label{eq:discrete_laplacian}
\end{align}
There are inherent discretization errors that we need to carefully evaluate. More on this will be addressed later in our analysis.

Consequently, to obtain the $O(\h^1)$ correction, we need to evaluate a nested integral. The pseudocode detailing our calculation method is presented in \cref{alg:1st_logqsolver}.
We use {\tt scipy.integrate.solve\verb|_|ivp} \cite{2020SciPy-NMeth} both in $\text{\tt 0th-logqSolver}$ and discrete-time update in $\text{\tt 1th-logqSolver}$.
We take $T_\text{min}>0$ as the initial time and $T_\text{max}$ as the terminal time for probability flow ODE\@.
This amounts to calculating the NLL for generation by SDE \eqref{eq:1para_gen_sde} in time interval $[T_\text{min}, T_\text{max}]$, and does not affect the sampling quality if we take sufficiently small $T_\text{min}$.
We show our results in \cref{tab:nll}.

\begin{algorithm}[t]
  \caption{$\text{\tt 1st-logqSolver}$}
  \label{alg:1st_logqsolver}
\begin{algorithmic}
  \STATE {\bfseries Input:} data $\mathbf x$, small $\Delta x$, SDE info $[{\bm f}(\cdot, t), g(t), T_\text{min}, T_\text{max}]$, solver $\text{\tt 0th-logqSolver}$  
  \STATE {\bfseries Initialize:} $t=T_\text{min}$, $\mathbf x_t = \mathbf x$, $\delta \mathbf x_t = \mathbf 0$, $\delta \log q_t = 0$
  \STATE Calculate $\log q_0 \leftarrow$ \text{\tt 0th-logqSolver}($\mathbf x, t=T_\text{min}$)
  \REPEAT
  \STATE Calculate 
  \begin{align*}
    \log q_t &\leftarrow \text{\tt 0th-logqSolver}(\mathbf x_t, t)
    \\
    \log q_t^{x+} &\leftarrow \text{\tt 0th-logqSolver}(\mathbf x_t + [\Delta x,0]^\top, t)
    \\
    \log q_t^{x-}&\leftarrow \text{\tt 0th-logqSolver}(\mathbf x_t - [\Delta x,0]^\top, t)
    \\
    \log q_t^{y+} &\leftarrow \text{\tt 0th-logqSolver}(\mathbf x_t + [0,\Delta x]^\top, t)
    \\
    \log q_t^{y-} &\leftarrow \text{\tt 0th-logqSolver}(\mathbf x_t - [0,\Delta x]^\top, t)
  \end{align*}
  \STATE Calculate $\bm f^{\rm PF}_{\theta} (\mathbf x_t,t)$
  \STATE Calculate $\delta {\bm f}^{\rm PF}_{\theta}(\mathbf x_t, \delta \mathbf x_t, t)$ by $\log q_t^{x/y\pm}$ and \eqref{eq:discrete_nabla}
  \STATE Calculate $\nabla \cdot \delta {\bm f}^{\rm PF}_{\theta}(\mathbf x_t, \delta \mathbf x_t, t)$ by $\log q_t^{x/y\pm}$ and \eqref{eq:discrete_laplacian}
  \STATE Update $t, \mathbf x_t, \delta \mathbf x_t, \delta \log q_t$ by time-discretized version (e.g. Runge-Kutta update) based on
  \begin{align*}
    \dot{\mathbf x}_t &= \bm f^{\rm PF}_{\theta} (\mathbf x_t,t)
    \\
    \dot{\delta \mathbf x}_t & = \delta {\bm f}^{\rm PF}_{\theta}(\mathbf x_t, \delta \mathbf x_t, t)
    \\
    \dot{\delta \log q_t} & = \nabla \cdot \delta {\bm f}^{\rm PF}_{\theta}(\mathbf x_t, \delta \mathbf x_t, t)
  \end{align*}
  \UNTIL{$t=T_\text{max}$}
  \STATE $\text{1st-correction} = \delta \mathbf x_T \cdot \nabla \log \pi (\mathbf x_T)
  + \delta \log q_T$
  \STATE {\bfseries Output:}  $\log q_0, \text{1st-correction}$
\end{algorithmic}
\end{algorithm}

\paragraph{Numerical errors}
In \cref{alg:1st_logqsolver}, we made two approximations in discrete differential operators and ODE solvers.
To ensure reliability of our results, 
it is essential to assess and estimate the associated errors.
Let us call the numerical value of $\log q^{\h=0}_t(\mathbf x)$ calculated by {\tt 0th-logqSolver} in \cref{alg:1st_logqsolver} as $N[\log q^{\h=0}_t(\mathbf x)]$, then we have two local errors:
\begin{equation}
  \small
  \begin{split}
    \text{err} \bigl(\hat{\nabla} N[\log q^{\h=0}_t(\mathbf x)]\bigr)
    &=
    \nabla \log q^{\h=0}_t(\mathbf x)
    \!-\!
    \hat{\nabla} N[\log q^{\h=0}_t(\mathbf x)]
    ,
    \\
    \text{err} \bigl(\hat{\nabla}^2 N[\log q^{\h=0}_t(\mathbf x)]\bigr)
    &=
    \nabla^2 \log q^{\h=0}_t(\mathbf x)
    \!-\!
    \hat{\nabla}^2 N[\log q^{\h=0}_t(\mathbf x)]
    .
    \label{eq:differential_errors}
  \end{split}
\end{equation}
Exact calculation of these values is unachievable, and nevertheless, we should somehow estimate them.
We propose two estimation schemes, {\tt subtraction} (\cref{app:sub}) and {\tt model} (\cref{app:model}).
Here, we show results based on the latter method, which operates at a higher speed.
In addition, we should integrate these local errors to estimate the error piled up in the final results.
This can also be calculated numerically by the ODE solver, and the numerical errors estimated in this way are shown in \cref{tab:nll}.
See \cref{app:errors} for more details.

\begin{figure}
    \centerline{\includegraphics[width=0.95\columnwidth]{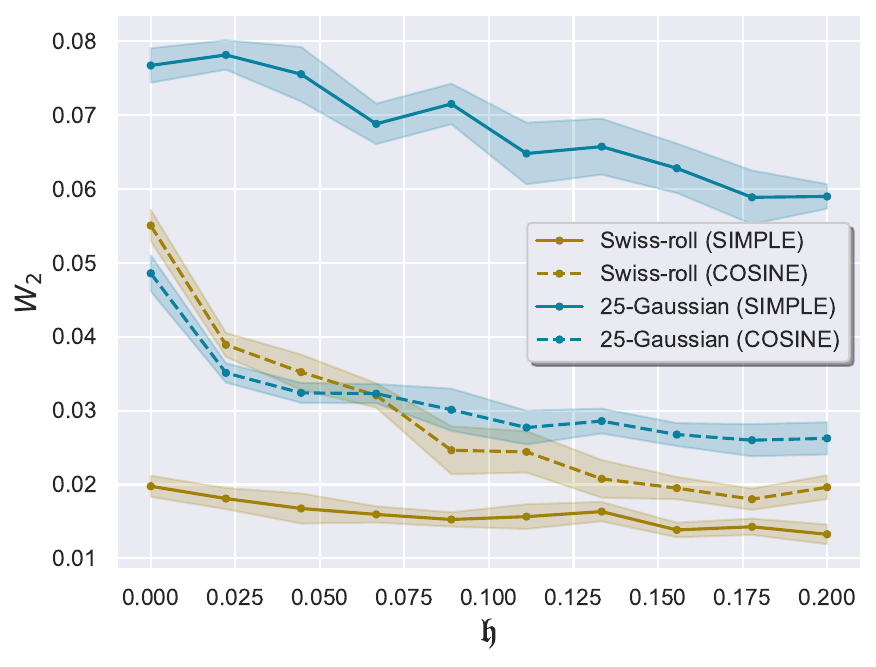}}
    \caption{2-Wasserstein metrics ($W_2$) by {\tt POT} library \cite{flamary2021pot} between validation data and generated data via \eqref{eq:1para_gen_sde} with Swiss-roll data and 25-Gaussian data with SIMPLE and COSINE SDE scheduling.  
    The dots and the errorbars represent the mean values over 10 independent trials and $\pm$std$/\sqrt{10}$, respectively.}
    \label{fig:w2}
\end{figure}

\paragraph{Comparison to 2-Wasserstein}
We also show 2-Wasserstein distance, $W_2$, between validation data and generated data in \cref{fig:w2} with stochastic SDE \eqref{eq:1para_gen_sde}.
We see that the $W_2$ values typically decrease especially in the small $\h$ region, which signifies the improvement of generated data quality.  This observation is consistent with negative-valued NLL corrections in \begin{sc}1st-corr\end{sc} column in \cref{tab:nll}.
The overall trend is the same as the results from the analytical study in \cref{sec:1d-gauss};
the tendency of enhancing sampled data quality by noise has been confirmed by our experiment even in intricate cases where exact calculation is not accessible.
%
In addition, considering that the values in \begin{sc}1st-corr\end{sc} column are the first-order derivative of NLL (cross-entropy) with respect to $\h$, one might say that the values for \begin{sc}cosine-SDE\end{sc} are
a few times larger than the values for \begin{sc}simple-SDE\end{sc}, which agrees with the behavior of the first-order derivative near $\h = 0$ in \cref{fig:w2}.

\begin{table}[t]
  \caption{NLL (cross-entropy) and its $O(\h^1)$ corrections. We apply $\Delta x$ = 0.01 to compute \eqref{eq:discrete_nabla} and \eqref{eq:discrete_laplacian}.
  {\tt tol} represents the value of absolute and relative tolerances of updates in \cref{alg:1st_logqsolver}.
  We set absolute and relative tolerances of {\tt 0th-logqSolver} as order {\tt 1e-5}.
  }
  \label{tab:nll}
  \begin{center}
  \begin{small}
  \begin{sc}
  \begin{tabular}{ccll}
    \multicolumn{4}{c}{Swiss-roll}\\
  \toprule
  SDE (NLL)
  & {\tt tol} & 1st-corr & errors \\
  \midrule
  \multirow{2}{*}{
    \begin{tabular}{c}
      simple
      \\
      (1.39$\pm$ 0.05)
    \end{tabular} 
  }
  &
  {\tt 1e-3}
  & -0.31$\pm$0.21
  & 0.13$\pm$0.00
  \\
  &
  {\tt 1e-5}
  & -0.44$\pm$0.38 & 
  0.13$\pm$0.00
  \\
  \midrule
  \multirow{2}{*}{
    \begin{tabular}{c}
      cosine
      \\
      (1.42$\pm$ 0.02)
    \end{tabular} 
  }
  &
  {\tt 1e-3}
  & -1.59$\pm$0.57
  & 0.35$\pm$0.00
  \\
  &
  {\tt 1e-5}
  & -3.27$\pm$1.11 & 
  0.37$\pm$0.02\\
  \bottomrule\\
  \multicolumn{4}{c}{25-gaussian}\\
      \toprule
      SDE (NLL)
      & {\tt tol} & 1st-corr & errors \\
      \midrule
      \multirow{2}{*}{
        \begin{tabular}{c}
          simple
          \\
          (-1.22$\pm$ 0.01)
        \end{tabular} 
      }
      &
      {\tt 1e-3}
      & -3.64$\pm$0.49
      & 0.31$\pm$0.00
      \\
      &
      {\tt 1e-5}
      & -3.61$\pm$0.64 & 
      0.32$\pm$0.01
      \\
      \midrule
      \multirow{2}{*}{
        \begin{tabular}{c}
          cosine
          \\
          (-1.71$\pm$ 0.02)
        \end{tabular} 
      }
      &
      {\tt 1e-3}
      & -17.57$\pm$5.56
      & 0.70$\pm$0.01
      \\
      &
      {\tt 1e-5}
      & -19.65$\pm$17.46 & 
      0.67$\pm$0.03\\
      \bottomrule
      \end{tabular}
  \end{sc}
  \end{small}
  \end{center}
\end{table}

\section{ Conclusions }


We presented a novel formulation of diffusion models utilizing the path integral framework, originally developed in quantum physics. 
This formulation provides a unified perspective on various aspects of score-based generative models, and we gave the re-derivation of reverse-time SDEs and loss functions for training. 
In particular, one can introduce a continuous parameter linking different sampling schemes: the probability flow ODEs and stochastic generation. 
We have performed the expansion with respect to this parameter and perturbatively evaluated the negative log-likelihood, which is a reminiscent of the WKB expansion in quantum physics. 
In this way, this formulation has presented a new method for scrutinizing the role of noise in the sampling process. 

%
An interesting future direction is an extension of the analysis based on path integral formalism for diffusion Schr\"{o}dinger bridges \cite{NEURIPS2021_940392f5}, in which the prior can be more general. 
Another direction is to understand the cases mentioned in \cite{karras2022elucidating}, where injecting noise would rather degrade the quality of the generated data. This is an open problem, and there is room for deeper study in terms of log-likelihood calculations based on WKB expansions.
%

%

\paragraph{Limitations} 
Our experiments did not involve actual image data. This omission is primarily due to the current evaluation method's limitations regarding NLLs, which are not directly applicable to high-dimensional data such as images on account of the uses of explicit discrete differentials.
%
%
Another limitation is the potential of underestimated numerical error in our computed NLLs. 
Although our estimated local errors look safe (\cref{app:vis_errors}), they are potentially underestimated because the estimation is based on score-based model that does not match to $\nabla \log q_t(\mathbf x)$ exactly in general.

\section*{Acknowledgements}
The work of Y.~H.~was supported in part by JSPS KAKENHI Grant No. JP22H05111.  
The work of A.~T.~was supported in part by JSPS KAKENHI Grant No. JP22H05116.
This work of K.~F.~was supported by JSPS KAKENHI Grant Nos. JP22H01216 and JP22H05118.




\bibliography{icml2024_ref}
\bibliographystyle{icml2024}

\newpage
\appendix
\onecolumn
\section{From SDE to path integral}
\label{app:to_path-integral}

In this appendix, we describe the details of the path integral formulation of diffusion models.

\subsection{ Discretization schemes }

In later discussions, we will introduce discretized summations in different ways, which do not seem apparent in their continuum counterpart.
We here introduce discretization schemes that will be used in later analyses. 

Suppose we have $\{\mathbf x_t \}_{t \in [0,T]}$ obeying SDE \eqref{eq:sde}. 
We discretize the time interval $[0,T]$ with width $\Delta t$ and $M \coloneqq T / \Delta t$, which we take to be an integer. 
Let us take the following summation: 
\begin{equation}
\sum_{n=0}^{M-1} 
\bm f_{t_n} \cdot (\mathbf x_{{t_n}+\Delta t} - \mathbf x_{t_n}) ,
\end{equation}
where $t_n \coloneqq n \Delta t$
and $\bm f_t = \bm f (\mathbf x_t, t)$.
We will consider the continuum limit $\Delta t \to 0$ and will refer to this 
as the It\^o scheme, which will be denoted by 
\begin{equation}
\int 
\bm f \cdot \rd \mathbf x 
\coloneqq 
\int \bm f_t \cdot (\mathbf x_{t+\rd t} - \mathbf x_t). 
\end{equation}
A common prescription in the physics literature 
is the Stratonovich scheme, 
\begin{equation}
\int 
\bm f \circ \rd \mathbf x 
\coloneqq 
\int 
\frac{
\bm f_t + \bm f_{t+\rd t}
}2 
\cdot 
(\mathbf x_{t+\rd t} - \mathbf x_t) ,
\end{equation}
which should be understood as a short-hand notation for the following equation,
\begin{equation}
\lim_{\Delta t \to 0} 
\sum_{n=0}^{M-1} 
\frac{
\bm f_{t_n} + \bm f_{t_n + \Delta t} 
}2 
\cdot 
(\mathbf x_{{t_n}+ \Delta t} - \mathbf x_{t_n}) .
\end{equation}
We also encounter the reverse-It\^o scheme, which is written as
\begin{equation}
\int 
\bm f \, \tilde{\cdot}\, \rd \mathbf x
\coloneqq 
\int \bm f (\mathbf x_{t+\rd t}) 
\cdot 
(\mathbf x_{t+\rd t} - \mathbf x_t) .
\end{equation}
This can be seen as the It\^o scheme in reverse time.

\subsection{ Derivation of the path integral expression }\label{app:deriv-path-int}

Here, we give a derivation of the path integral representation of a diffusion model (\cref{thm:path-integral}).
We consider a forward process described by the SDE \eqref{eq:sde}.
In the physics literature, 
Eq.~\eqref{eq:sde} is commonly expressed in the following form, 
\begin{equation}
\quad \dot{ \mathbf x }_t 
= \bm f (\mathbf{x}_t, t)  +  \bm \xi_t ,
\label{eq:eom-2}
\end{equation}
where the noise term satisfies
\begin{equation}
\mathbb E[
\xi^i_t  \xi^j_{t'}
]
= g(t)^2 \delta^{ij} \delta(t - t'). 
\end{equation}
The expectation value of a generic observable $\mathcal O(\{\mathbf x_t \})$ 
over noise realizations can be expressed as 
\begin{equation}
\mathbb E_{P[\{\mathbf x_t\}]}
[ \mathcal O (\{\mathbf x_t\}) ]
= 
\int [D \mathbf x_t] [D \bm \xi_t]
\, 
\mathcal O (\{ \mathbf x_t \})
\left(
\prod_t \delta (\mathbf x_t - \mathbf x^{\rm sol}_t)
\right)
e^{- \int_0^T  \frac{1}{2g(t)^2} 
\| \bm \xi_t\|^2 \rd t } 
p_0 (\mathbf x_0), 
\label{eq:path-int-average}
\end{equation}
where $p_0 (\mathbf x_0)$ is the probability distribution of initial states (i.e., the data distribution), 
$\delta(\cdot)$ is Dirac's delta function,
and $\mathbf x^{\rm sol}_t$ is the solution of Eq.~\eqref{eq:eom-2}.
The symbol $[D \mathbf x_t]_{\Delta t}$ 
denotes $[D\mathbf x_t]_{\Delta t} \coloneqq \prod_{n=0}^{M} 
C(\Delta t)
\rd \mathbf x_{t_n}
$, where $C(\Delta t)$ is a normalization constant,
and similarly for $[D \bm \xi_t]$. 
When the weight is sufficiently well-behaved 
(like $e^{- \int_0^T  \frac{1}{2g(t)^2} 
\| \bm \xi_t\|^2 \rd t }$ in \cref{eq:path-int-average}), 
the limit $\Delta t \to 0$
the path integral is well-defined.
We will be considering this limit at the end of the calculation do not explicitly indicate
the dependence on $\Delta t$ hereafter. 
The delta function imposes that $\mathbf x_t$ 
is a solution of Eq.~\eqref{eq:eom-2} under a given noise realization.
Using the change-of-variable formula for the 
delta function, 
\begin{equation}
\delta (\mathbf x_t - \mathbf x^{\rm sol}_t)
= 
\left| 
 \det \frac{\delta {\rm EOM}_t}{\delta \mathbf x_{t'}}
 \right|
 \delta ({\rm EOM}_t) ,
\end{equation}
where ${\rm EOM}_t \coloneqq \dot{\mathbf  x}_t - \bm f (\mathbf{x}_t, t) -  \bm \xi_t $.
The Jacobian part gives a nontrivial contribution (see \cref{sec:jacobian} for a detailed derivation)
in the case of the Stratonovich scheme, 
\begin{equation}
\begin{split}
\left| 
 \det \frac{\delta {\rm EOM}_t}{\delta \mathbf x_{t'}}
 \right|
&= 
\det 
( \p_t \delta_{ij} - \p_i f_j )
\\
&= 
\det \p_t
\, 
\det 
(
\delta_{ij} - \p_t^{-1} \p_i f_j
) 
\\
&\propto 
e^{- \frac{1}2 
\int \nabla \cdot \bm f(\mathbf{x}_t, t) \rd t
}, 
\end{split}
\end{equation}
while it gives a trivial factor for the It\^o scheme.
Writing this factor as $e^{-J}$, 
the expectation value can now be written as 
\begin{equation}
\mathbb E_{P[\{\mathbf x_t\}]}
[ \mathcal O (\{ \mathbf x_t \}) ]
= 
\int [D \mathbf x_t][D \bm \xi_t]
\, 
\mathcal O (\{ \mathbf x_t \})
e^{-J} 
\left( 
\prod_t
\delta (\dot{\mathbf x}_t - \bm f_t - \bm \xi_t)
\right)
e^{- \int_0^T  \frac{1}{2g(t)^2} 
\| \bm \xi_t\|^2 \rd t }
p_0 (\mathbf x_0) .  
\end{equation}
Performing the integration over $\bm \xi_t$, 
we arrive at the expression
\begin{equation}
\mathbb E_{P[\{\mathbf x_t\}]}
[\mathcal O (\{ \mathbf x_t \})]
= 
\int [D \mathbf x_t]
\, 
\mathcal O (\{ \mathbf x_t \})
\, 
e^{ - \Acal}  
p_0 (\mathbf x_0),
\label{eq:o-exp}
\end{equation}
Here, we defined the action $\Acal$ by 
\begin{equation}
\Acal \coloneqq \int_0^T  L (\dot{\mathbf{x}}_{t}, \mathbf{x}_{t}) \rd t + J
\end{equation}
where $L(\dot{\mathbf{x}}_{t}, \mathbf{x}_{t})$ is the Onsager-Machlup function 
~\cite{PhysRev.91.1505,PhysRev.91.1512}
given by
\begin{equation}
L(\dot{\mathbf{x}}_{t}, \mathbf{x}_{t}) =
\frac{1}{2g(t)^2} 
\| \dot{\mathbf x}_t - \bm f (\mathbf{x}_t, t) \|^2,
\end{equation}
and 
$J$ is the term coming from the Jacobian, which depends on the choice of the discretization scheme:
\begin{equation}
J
= 
\begin{cases}
0 & \text{It\^o} 
\\
\int_0^T 
\frac{1}2 \nabla \cdot \bm f_t (\mathbf{x}_t, t)
\rd t
& \text{Stratonovich}
\\
\int_0^T 
 \nabla \cdot \bm f_t (\mathbf{x}_t, t)
\rd t
& \text{Reverse-It\^o} 
\end{cases}. 
\end{equation}
This concludes the derivation of \cref{thm:path-integral}. 
Any observable can be computed using the weight 
given by $\Acal$. 
For example, one can compute the joint probability 
for the variables on discrete time points 
$\{t_n\}_{n = 1, \ldots, N}$ as
\begin{equation}
P(\mathbf y_{t_1}, \mathbf y_{t_2}, \ldots, \mathbf y_{t_N}) 
= 
\int [D \mathbf x_t]
\, 
e^{ - \Acal} \, p_0 (\mathbf x_0) 
\,
\prod_{n=1}^N \delta (\mathbf y_{t_n} - \mathbf x_{t_n}) . 
\end{equation}

\subsection{ Time-reversed SDE } \label{app:derivation-reverse-sde}

Here we present the derivation of the time-reversed dynamics in the path integral formalism (\cref{thm:reverse_sde}).
We will use the It\^o scheme, and later comment on other discretization schemes.
We start by rewriting Eq.~\eqref{eq:o-exp} as
\begin{equation}
\begin{split}
\mathbb E_{P[\{\mathbf x_t\}]}
[\mathcal O (\{ \mathbf x_t \})]
&= 
\int [D \mathbf x_t ]
\, 
\mathcal O (\{\mathbf x_t\}) 
\, 
p_T (\mathbf x_T) 
\,
e^{\int_0^T 
 [ - L - \ln p_T (\mathbf x_T) + \ln p_0 (\mathbf x_0) ]\rd t 
} 
\\
&= 
\int [D \mathbf x_t ]
\, 
p_T (\mathbf x_T)
\, 
e^{- \wt{\Acal} 
} ,
\end{split}
\end{equation}
where 
$\wt{\Acal}
\coloneqq \int_0^T  \wt{L} (\dot{\mathbf{x}}_{t}, \mathbf{x}_{t}) \rd t 
$ with 
\begin{equation}
\wt{L} 
(\dot{\mathbf{x}}_{t}, \mathbf{x}_{t})
\coloneqq L 
(\dot{\mathbf{x}}_{t}, \mathbf{x}_{t})
+ \frac{\rd}{\rd t} \ln p_t (\mathbf x_t) . 
\end{equation}
The total time derivative of $\ln p_t(\mathbf x_t)$ is written as 
\begin{equation}
\rd  \ln p_t (\mathbf x_t) 
 = 
 \p_t 
 \ln p_t (\mathbf x_t) \rd t 
 + 
\rd \mathbf x_t \cdot  \nabla \ln p_t (\mathbf x_t)
 + 
 \frac{1}2 g(t)^2
 \nabla^2 \ln p_t (\mathbf x_t) \rd t ,
 \label{eq:dp-ito}
\end{equation}
where we have used It\^o's formula.
For $\p_t p_t (\mathbf x_t)$, we use the Fokker-Planck equation,
\begin{equation}
\p_t p_t (\mathbf x) =
- \nabla \cdot 
\left( 
\bm{f}(\mathbf{x}_t, t)
p_t (\mathbf x) - \frac{g(t)^2}{2} \nabla p_t (\mathbf x)
\right),
\end{equation}
which we rewrite as 
\begin{equation}
\p_t \ln p_t (\mathbf x)    
= 
- \nabla \cdot \bm{f}(\mathbf{x}_t, t)
- \bm{f}(\mathbf{x}_t, t) \cdot \nabla \ln p_t (\mathbf x) 
+ 
\frac{g(t)^2}2
\left[
\nabla^2 \ln p_t (\mathbf x) + (\nabla \ln p_t (\mathbf x) )^2 
\right] .
\end{equation}
The new Lagrangian $\wt{L}(\dot{\mathbf{x}}_{t}, \mathbf{x}_{t})$ is now written as 
\begin{equation}
\begin{split}
\wt{L} (\dot{\mathbf{x}}_{t}, \mathbf{x}_{t})
&= 
\frac{1}{2g_t^2} 
(
\dot{\mathbf x}^2  -
2 \dot{\mathbf x} \cdot 
(
\bm{f}(\mathbf{x}_t, t)
- g(t)^2 \nabla \ln p_t (\mathbf x_t) 
)
+ 
\bm{f}(\mathbf{x}_t, t)^2
)
+ 
\frac{g(t)^2}2  \nabla^2 \ln p_t (\mathbf x_t) 
\\
&\quad 
- \nabla \cdot \bm{f}(\mathbf{x}_t, t)
- 
\bm{f}(\mathbf{x}_t, t)
\cdot \nabla \ln p_t (\mathbf x_t) 
+ \frac{g(t)^2}2 
\left[
\nabla^2 \ln p_t (\mathbf x_t) + (\nabla \ln p_t (\mathbf x_t) )^2 
\right]
\\
&= 
\frac{1}{2g_t^2} 
(
\dot{\mathbf x} - 
\bm{f}(\mathbf{x}_t, t)
+ g(t)^2 \nabla \ln p_t (\mathbf x_t) 
)^2
- \nabla \cdot 
\left( 
\bm{f}(\mathbf{x}_t, t)
- g(t)^2 \nabla \ln p_t (\mathbf x_t)
\right). 
\end{split}
\end{equation}
Thus, the action $\wt{\mathcal A}$ 
is given by 
\begin{equation}
\int_0^T 
 \wt{L} (\dot{\mathbf{x}}_{t}, \mathbf{x}_{t}) \rd t
= 
\int_0^T 
\left[ 
\frac{1}{2g(t)^2} 
\| 
\dot{\mathbf x}_t - 
\wt{\bm f}(\mathbf{x}_t, t)
\|^2 
- \nabla \cdot \wt{\bm f}(\mathbf{x}_t, t)
\right] 
\rd t
,
\label{eq:lp-ito}
\end{equation}
where
\begin{equation}
\wt{\bm f} (\mathbf{x}_t, t)
\coloneqq 
\bm f (\mathbf{x}_t, t)
- g(t)^2 \nabla \ln p_t (\mathbf x_t) . 
\end{equation}
Note that there appears a nontrivial Jacobian 
term in Eq.~\eqref{eq:lp-ito}. 
This term disappears if we rewrite the integral 
using the inverse time $\tau$ in the It\^o convention. 
The action $\wt{\Acal}$ contains the following contribution,
\begin{equation}
\wt{\Acal}
\supset 
- \int \frac{1}{g(t)^2} \wt{\bm f}(\mathbf{x}_t, t) \cdot \rd \mathbf x_t .
\end{equation}
Currently, this product written in the It\^o scheme.
We rewrite this term in the reverse-It\^o scheme,
\begin{equation}
\begin{split}    
\wt{\bm f}_t  \cdot \rd \mathbf x_t 
&= 
\left[
\wt{\bm f}_{t+\rd t} 
- 
\left( 
\wt{\bm f}_{t+\rd t} 
- 
\wt{\bm f}_{t} 
\right)
\right]
\cdot \rd \mathbf x_t 
= 
\wt{\bm f}_{t+\rd t} 
\cdot \rd \mathbf x_t 
- 
\sum_{i,j} \p_i (\wt{f}_t)_j \rd x_i \rd x_j
\\
&= 
\wt{\bm f}_{t+\rd t} 
\cdot \rd \mathbf x_t 
- 
g(t)^2  \nabla \cdot \wt{\bm f}_{t} \, \rd  t . 
\end{split}
\end{equation}
The second term of this equation cancels the Jacobian term. 
Thus, the time-reversed action can be naturally interpreted as an It\^o integral in reverse time.

The same procedure can be also done 
for the Stratonovich scheme.
The difference is the presence of the Jacobian term in the original action and 
the total time derivative of $p_t (\mathbf x_t)$ is written as 
\begin{equation}
 \rd 
 \ln p_t (\mathbf x_t) 
 = 
 \p_t
 \ln p_t (\mathbf x_t)
 \rd t 
 + 
\rd \mathbf x_t \cdot 
 \nabla \ln p_t (\mathbf x_t)
\end{equation}
instead of Eq.~\eqref{eq:dp-ito}. 
Following similar steps, the time-reversed action is found to be given by
%
%
%
\begin{equation}
\wt{\Acal}
= 
\int_0^T 
 \wt{L} (\dot{\mathbf{x}}_{t}, \mathbf{x}_{t}) \rd t
= 
\int_0^T 
\left[ 
\frac{1}{2g(t)^2} 
\|
\dot{\mathbf x}_t - \wt{\bm f}_t
\|^2 
- \frac{1}2 \nabla \cdot \wt{\bm f}_t
\right] 
\rd t
.
\end{equation}
Note that the sign of the Jacobian term is flipped compared with the original action,
which allows us to interpret the process as a time-reversed one.



\subsection{ Evaluation of KL divergence }\label{app:eval-kl}

We here give the derivation of \cref{thm:KL}. 
We evaluate the upper limit of the KL divergence of the data distribution and a model, 
\begin{equation}
D_{\rm KL} (p_0(\mathbf x_0) \| q_0(\mathbf x_0) )
\le 
D_{\rm KL} (
P(\{ \mathbf x_t \}_{t \in [0,T]}) 
\| 
Q(\{ \mathbf x_t \}_{t \in [0,T]}) 
),
\label{eq:data-proc}
\end{equation}
where we used the data processing inequality. 
Below, we evaluate the RHS of \cref{eq:data-proc}. 

The time-reversed action of the data distribution (in the reverse-It\^o scheme) reads 
\begin{equation}
\wt{\Acal}
\coloneqq 
\int_0^T 
 \wt{L}(\dot{\mathbf{x}}_{t}, \mathbf{x}_{t})\rd t
= 
\int_0^T 
\left[ 
\frac{1}{2g(t)^2} 
\|
\dot{\mathbf x}_t - \wt{\bm f} (\mathbf{x}_t, t)
\|^2 
\right] 
\rd t
,
\label{eq:a-tilde-rev}
\end{equation}
The time-reversed action of a model is given by
\begin{equation}
\wt{\Acal}_\theta
\coloneqq 
\int_0^T 
 \wt{L}_\theta (\dot{\mathbf{x}}_{t}, \mathbf{x}_{t}) \rd t
= 
\int_0^T 
\left[ 
\frac{1}{2g(t)^2} 
\|
\dot{\mathbf x}_t - \wt{\bm f}_{\theta} (\mathbf{x}_t, t)
\|^2 
\right] 
\rd t
,
\label{eq:a-tilde-model-rev}
\end{equation}
with 
\begin{equation}
\wt{\bm f}_{\theta} (\mathbf{x}_t, t)
\coloneqq     
\bm f(\mathbf{x}_t, t) - g(t)^2 \bm{s}_\theta (\mathbf x_t, t) .
\end{equation}
We note that, we employ the reverse-It\^o scheme
in the following computation 
and that is why there is no Jacobian term 
in \cref{eq:a-tilde-rev,eq:a-tilde-model-rev}.
We rewrite the joint probability of paths as
\begin{align}    
P(\{ \mathbf x_t \}_{t \in [0,T]}) 
&= 
e^{- \Acal} p_0(\mathbf x_0 )
= 
p_T (\mathbf x_T) e^{- \wt{\Acal}} ,
\\
Q(\{ \mathbf x_t \}_{t \in [0,T]}) 
&= 
e^{- \Acal_\theta} q_0(\mathbf x_0 )
= 
q_T (\mathbf x_T) e^{- \wt{\Acal}_\theta} . 
\end{align}
The KL divergence of 
the path-probability 
$P(\{\mathbf x_t\}_t)$ from 
$Q(\{\mathbf x_t\}_t)$ is written as
\begin{equation}
\begin{split}
D_{\rm KL} \left( 
P(\{ \mathbf x_t \}_{t \in [0,T]})
\| 
Q(\{ \mathbf x_t \}_{t \in [0,T]})
\right)
&=
\mathbb E_{P(\{ \mathbf x_t \})} 
\left[ 
\ln 
\frac{ P(\{ \mathbf x_t \}_{t \in [0,T]}) }{Q(\{ \mathbf x_t \}_{t \in [0,T]})}
\right]
\\
&= 
\mathbb E_{P(\{ \mathbf x_t \}_t)} 
\left[ 
\ln \frac{ p_T(\mathbf x_T) }{ q_T(\mathbf x_T) } -\wt{ \Acal} + \wt{\Acal}_\theta
\right] 
\\
&= 
D_{\rm KL} \left( p_T (\mathbf x_T) \| q_T(\mathbf x_T) \right)
+ 
\mathbb E_{P(\{ \mathbf x_t \}_t)} 
\left[ \wt{\Acal}_\theta - \wt{\Acal} \right] .
\end{split}
\end{equation}
The second term of the RHS can be written as
\begin{equation}
\begin{split}
\mathbb E_{P(\{ \mathbf x_t \}_t)} 
\left[ \wt{\Acal}_\theta - \wt{\Acal} \right] 
&= 
\mathbb E_{P(\{ \mathbf x_t \}_t)} 
\int_0^T 
\frac{1}{2g(t)^2} 
\left[ 
- 2 \dot{\mathbf x}_t \tilde{\cdot} ( \wt{\bm f}_\theta (\mathbf{x}_t, t) - \wt{\bm f} (\mathbf{x}_t, t)) 
+ (\wt{\bm f }_\theta (\mathbf{x}_t, t))^2 - (\wt{\bm f} (\mathbf{x}_t, t))^2 
\right]
\rd t
\\ 
&= 
\mathbb E_{P(\{ \mathbf x_t \}_t)} 
\int_0^T 
\frac{1}{2g(t)^2} 
[\wt{\bm f}_\theta (\mathbf{x}_{t+\rd t}, t+\rd t) - \wt{\bm f} (\mathbf{x}_{t+\rd t}, t+\rd t)] \cdot 
[
- 2 \dot{\mathbf x}_t + \wt{\bm f}_\theta (\mathbf{x}_t, t) + \wt{\bm f} (\mathbf{x}_t, t)
]
\rd t
. 
\end{split}
\end{equation}
We discretize this and look at the contribution from 
the neighboring part $(t, t+\Delta t)$, 
\begin{equation}
\begin{split}
&
\mathbb E_{P(\mathbf x_t, \mathbf x_{t+\Delta t})}
\left[ 
\Delta t 
\, 
\frac{1}{2g(t)^2} 
(\wt{\bm f}_{\theta,t+\Delta t} - \wt{\bm f}_{t+\Delta t}) \cdot
\left[ 
- 2 (\mathbf x_{t + \Delta t} - \mathbf x_t) / \Delta t 
+ \wt{\bm f}_{\theta,t} + \wt{\bm f}_t
\right]
\right] 
\\
&=     
\mathbb E_{p_t(\mathbf x_t)}
\left[ 
\Delta t 
\, 
\frac{1}{2g(t)^2} 
(\wt{\bm f}_{\theta,t+\Delta t} - \wt{\bm f}_{t+\Delta t}) 
\cdot
( 
- 2 \wt{\bm f}_t 
+ \wt{\bm f}_{\theta,t} + \wt{\bm f}_t
)
\right]
\\
&= 
\mathbb E_{p_t(\mathbf x_t)}
\left[ 
\Delta t 
\, 
\frac{1}{2g(t)^2} 
\| \wt{\bm f}_{\theta,t} - \wt{\bm f}_t \|^2 
+ O(\Delta t^2)
\right] ,
\end{split}
\end{equation}
where we performed the summation over 
$\mathbf \delta = \mathbf x_{t+\Delta t} - \mathbf x_t$.
Summing up these contributions for $[0,T]$, we have 
\begin{equation}
\begin{split}
\mathbb E_{P(\{ \mathbf x_t \}_t)} 
\left[\wt{\Acal}_\theta - \wt{\Acal} \right] 
&= 
\mathbb E_{P(\{ \mathbf x_t \}_t)} 
\left[ 
\int_0^T  
\frac{1}{2g(t)^2} 
\| \wt{\bm f}_{\theta}(\mathbf{x}_t, t) - \wt{\bm f} (\mathbf{x}_t, t) \|^2 
\rd t
\right] 
\\
&=
\mathbb E_{P(\{ \mathbf x_t \}_t)} 
\left[ 
\int_0^T  
\frac{g(t)^2}{2} 
\|\nabla \ln p_t (\mathbf x_t) - \bm{s}_\theta (\mathbf{x}_t, t) \|^2 
\rd t
\right] 
. 
\end{split}
\end{equation}
Thus, we have obtained the following inequality, 
\begin{equation}
D_{\rm KL} (p_0(\mathbf x_0) \| q_0(\mathbf x_0) )
\le 
D_{\rm KL} \left( p_T (\mathbf x_T) \| q_T(\mathbf x_T) \right)
+ 
\int_0^T  
\frac{g(t)^2}{2} 
\mathbb E_{p_t}
\left[ 
\| \nabla \ln p_t (\mathbf x_t) - \bm{s}_\theta (\mathbf x_t, t) \|^2 
\right] 
\rd t
. 
\label{eq:kl-bound}
\end{equation}
The distribution $q_T(\mathbf x_T)$ is taken to be a prior, $\pi (\mathbf x_T)$. 
This concludes the proof 
of \cref{thm:KL}.

\subsection{Conditional variants}\label{app:conditional}

A similar argument applies to the case 
when the initial state is fixed.
The derivation of the reverse process 
can be obtained by replacing the probability 
densities with conditional ones on 
a chosen initial state $\mathbf x'_0$.
The expectation value of a general observable in this situation can be written as 
\begin{equation}
\mathbb E_{P(\{\mathbf x_t\}_{t \in (0,T]}|\mathbf x'_0)}
[ 
\mathcal O (\{ \mathbf x_t \})
]
= 
\int 
[D \mathbf x_t]
\, 
\mathcal O (\{ \mathbf x_t \})
\, 
e^{ - \int  L \rd t} 
\delta (\mathbf x_0 - \mathbf x'_0) .
\end{equation}
Similarly to the previous section, 
we can rewrite this as 
\begin{equation}
\begin{split}
\mathbb E_{P(\{\mathbf x_t\}_{t \in (0,T]}|\mathbf x'_0)}
[ 
\mathcal O (\{ \mathbf x_t \})
]
&= 
\int [D \mathbf x_t]
\, 
\mathcal O (\{x_t\}) 
\, 
P_T (\mathbf x_T | \mathbf x'_0 ) 
\,
e^{ - \wt{\Acal} (\mathbf x'_0) } 
\delta (\mathbf x_0 - \mathbf x'_0) 
, 
\end{split}
\end{equation}
where 
$
\wt{\Acal} (\mathbf x'_0)
\coloneqq 
\int_0^T 
\wt{L} (\dot{\mathbf x}_t, \mathbf x_t | \mathbf x'_0) 
\, \rd t 
$ 
with 
\begin{equation}
\wt{L} (\dot{\mathbf x}_t, \mathbf x_t | \mathbf x'_0) 
\coloneqq L (\dot{\mathbf x}_t, \mathbf x_t | \mathbf x'_0) 
+ \frac{\rd}{\rd t} p_t (\mathbf x_t | \mathbf x'_0 ).
\end{equation}
Following similar calculations with the unconditional case, the force of the reverse process turns out to be given by
\begin{equation}
\wt{\bm f}(\mathbf{x}_t, t)
\coloneqq 
\bm f (\mathbf{x}_t, t)
- g(t)^2 \nabla \ln p_t (\mathbf x_t | \mathbf x'_0) .
\label{eq:conditional_reverse_drift}
\end{equation}

One can also repeat a similar argument for 
the evaluation of the KL divergence 
with fixed $\mathbf{x}_0'$, 
which corresponds to the ELBO-based loss \cite{kingma2021variational, kingma2023understanding}: 
\begin{proposition}
  \label{thm:reverse_sde_cond}
  Let $p_t(\mathbf{x}|\mathbf{x}_0)$ as the Markov kernel from time $0$ to $t$ determined by the Fokker-Planck equation.
  Determining the Onsager–Machlup function for reverse process $\widetilde{L}(\dot{\mathbf{x}}_{t}, \mathbf{x}_{t}|\mathbf{x}_0')$ to satisfy
  \begin{align}
    &p_0(\mathbf{x}_0|\mathbf{x}_0') e^{-\int_0^T L(\dot{\mathbf{x}}_{t}, \mathbf{x}_{t}) \rd t + J} [D\mathbf{x}_t]
    = 
    e^{-\int_0^T \widetilde{L}(\dot{\mathbf{x}}_{t}, \mathbf{x}_{t}|\mathbf{x}_0') \rd t + \widetilde{J}(\mathbf{x}_0')} p_T(\mathbf{x}_T|\mathbf{x}_0') [D\mathbf{x}_t],
  \end{align}
  yields
  \begin{align}
    \widetilde{L}(\dot{\mathbf{x}}_{t}, \mathbf{x}_{t}|\mathbf{x}_0')
    &=
    \frac{\| \dot{\mathbf{x}}_t - \bm{f}(\mathbf{x}_t, t) + g(t)^2 \nabla \log p_t(\mathbf{x}_t|\mathbf{x}_0')\|^2}{2g(t)^2 }
    ,
  \end{align}
  where $\widetilde{J}(\mathbf{x}_0')$ is the conditional Jacobian for the reverse process depending on discretization scheme.
\end{proposition}
This representation proves to be practically valuable, especially since in most cases, we do not have access to the score function $\nabla \log p_t(\mathbf{x})$.

\begin{proposition}
  \label{thm:cond_KL}
  The KL divergence of path-probabilities can be represented by the path integral form,
    \begin{align}
    D_{\rm KL}(P(\cdot|\mathbf{x}_0')\|Q_\theta)
    &=
    \int 
    e^{- \wt \Acal (\mathbf x'_0)}
    p_T(\mathbf{x}_T)
    \log \frac{
      e^{- \wt \Acal (\mathbf x'_0) 
      }\,
      p_T(\mathbf{x}_T)
    }{
      e^{- \wt{\Acal}_\theta}
      \pi(\mathbf{x}_T)
    }[D\mathbf{x}_t] ,
  \end{align}
where $\pi$ is a prior distribution, 
and it can be computed as
\begin{equation}    
D_{\rm KL}(P(\cdot|\mathbf{x}_0')\|Q_\theta)
=
D_{\rm KL} ( p_T \| \pi )
+
\int_0^T \frac{g(t)^2}{2}
\mathbb{E}_{p_t}
\left[
 \| \nabla \log p_t(\mathbf{x}_t|\mathbf{x}_0') - \bm{s}_\theta(\mathbf{x}_t, t)\|^2
\right] 
\rd t .
\label{eq:loss_conditioned}
\end{equation}
\end{proposition}


\subsection{ Likelihood evaluation }\label{app:proof-q-h-formula}

We here give the proof of \cref{thm:likelihood_wkb}.

The probability distribution $q_t^\h (\mathbf x)$
corresponding to the backward process \eqref{eq:rescoresde} 
satisfies the following 
Fokker-Planck equation,
\begin{equation}
\p_t q^\h_t (\mathbf x)    
= 
- \nabla \cdot 
\left[ 
\bm f^{\rm PF}_{\theta} (\mathbf x, t)
q^\h_t (\mathbf x)
- 
\frac{\h g(t)^2}{2}
(\bm s_{\theta,t} ( \mathbf x) - \nabla \log q^\h_t(\mathbf x)) q^\h_t(\mathbf x)
\right] .
\label{eq:q-h-fp}
\end{equation}
Note that $q^\h_t(\mathbf x)$ depends on $\h$ nontrivially
when $\bm s_{\theta} (\mathbf x,t) \neq \nabla \log q^\h_t(\mathbf x)$. 
Introducing a new parameter $\gamma \in \mathbb R_{\ge 0}$, \cref{eq:q-h-fp} can be written as 
\begin{equation}
\begin{split}
\p_t q^\h_t (\mathbf x)    
&= 
- \nabla \cdot 
\left[ 
\bm f^{\rm PF}_{\theta} (\mathbf x,t)
q^\h_t (\mathbf x)
- 
\frac{\h g(t)^2}{2}
(\bm s_{\theta,t} ( \mathbf x) - \nabla \log q^\h_t(\mathbf x)) q^\h_t(\mathbf x)
- \frac{\gamma g(t)^2}2 \nabla q^\h_t (\mathbf x) 
+ 
\frac{\gamma g(t)^2}2 \nabla q^\h_t (\mathbf x) 
\right] 
\\
&= 
- \nabla \cdot 
\left[ 
\left( 
\bm f^{\rm PF}_{\theta} (\mathbf x,t)
- 
\frac{g(t)^2}{2}
\left[
\h (\bm s_{\theta,t} ( \mathbf x) - \nabla \log q^\h_t(\mathbf x))
+ \gamma \nabla \log q^\h_t(\mathbf x)
\right]
\right) 
q^\h_t(\mathbf x)
+ 
\frac{\gamma g(t)^2}2 \nabla q^\h_t (\mathbf x) 
\right]  .
\end{split}
\label{eq:fp-model}
\end{equation}
Let us define 
\begin{equation}
\bm F_{\h, \gamma} (\mathbf x, t)
\coloneqq 
\bm f^{\rm PF}_{\theta} (\mathbf x,t)
- 
\frac{\h g(t)^2}{2} 
\bm s_{\theta}(\mathbf x, t)
+ 
(\h - \gamma)
\frac{g(t)^2}{2}
\nabla \log q^\h_t (\mathbf x) .
\end{equation}
We consider the path integral expression 
for Eq.~\eqref{eq:fp-model}. 
The corresponding action is given by 
\begin{equation}
\wt{\Acal}_{\h,\gamma} 
= 
\int_0^T \rd t\,
\wt{L}_{\h,\gamma} (\dot{\mathbf{x}}_{t}, \mathbf{x}_{t})
= 
\int_0^T  
\frac{1}{2\gamma g(t)^2}
\| \dot{\mathbf x} - \bm F_{\h,\gamma} (\mathbf x_t, t)
\|^2 \, 
\rd t
.
\end{equation}
Using the It\^o scheme,
one can obtain the likelihood for finite $\h$ 
by taking the limit $\gamma \to 0$,
\begin{equation}
\begin{split}    
q^\h_0 (\mathbf x'_0 )
&
\propto 
\int 
\left(
\prod_{t \in [0,T]} \rd \mathbf x_t  
\right)
\, 
\delta (\mathbf x_0 - \mathbf x_0') 
e^{- \wt{\Acal}_{\h,\gamma} + \int_0^T \nabla \cdot \bm F_{\h,\gamma} \rd t} q^\h_T(\mathbf x_T)
\\
& \xrightarrow{\gamma \to 0}
e^{\int_0^T \nabla \cdot \bm F_{\h,\gamma =0} \rd t} q^\h_T(\mathbf x_T) . 
\end{split}
\end{equation}
Thus, we obtain the formula for the likelihood 
for finite $\h$,
\begin{equation}
\log q^\h_0 (\mathbf x_0) 
= 
\log q^\h_T (\mathbf x_T) 
+ 
\int_0^T 
\, 
\nabla \cdot
\left( 
\bm f^{\rm PF}_{\theta} (\mathbf x_t,t)
- 
\frac{\h g(t)^2}{2}
\left[
\bm s_{\theta} ( \mathbf x_t, t) - \nabla \log q^\h_t(\mathbf x_t)
\right]
\right)
\rd t
. 
\end{equation}
Since $q^\h_T (\mathbf x_T)$ is taken 
to be a prior, 
$q^\h_T (\mathbf x_T)  =\pi (\mathbf x_T)$, 
this concludes the proof of \cref{thm:likelihood_wkb}.

\subsection{ Likelihood up to first order of $\h$}\label{app:likelihood_perturbative}
We here give a proof of \cref{thm:likelihood_perturbative}.

First, we rename the solution for the modified probability flow ODE \eqref{eq:modified-p-flow} as $\mathbf x^\h_t$, i.e.
\begin{equation}
\begin{split}
  \dot{\mathbf x}^{\h}_t
  =
  \bm f^{\rm PF}_{\theta} (\mathbf x^\h_t,t)
  -
  \frac{\h g(t)^2}{2}
  \left[
  \bm s_{\theta} ( \mathbf x^\h_t, t) - \nabla \log q^\h_t(\mathbf x^\h_t)
  \right]
,
\quad 
\mathbf x^\h_{t=0} = \mathbf x_0
\end{split}
\end{equation}
and it can be represented by the formal integral
\begin{equation}
\begin{split}
  \mathbf x^\h_t
  =
  \mathbf x_0
  +
  \int_0^t
  \, 
  \left(
    \bm f^{\rm PF}_{\theta} (\mathbf x^\h_t,t)
    -
    \frac{\h g(t)^2}{2}
    \left[
    \bm s_{\theta} ( \mathbf x^\h_t, t) - \nabla \log q^\h_t(\mathbf x^\h_t)
    \right]
  \right)
  \rd t
  .
\end{split}
\label{eq:formal_xht}
\end{equation}
Next, we consider Taylor expansion of $\bm f^{\rm PF}_{\theta} (\mathbf x^\h_t,t)$ with $\h$:
\begin{equation}
\begin{split}
  \bm f^{\rm PF}_{\theta} (\mathbf x^\h_t,t)
  &=
  \bm f^{\rm PF}_{\theta} (\mathbf x_t,t)
  +
  \h \underbrace{
    \partial_\h\bm f^{\rm PF}_{\theta} (\mathbf x_t,t)|_{\h=0} 
  }_{
    (\partial_\h \mathbf x^\h_t|_{\h=0} \cdot \nabla) \bm f^{\rm PF}_{\theta} (\mathbf x_t,t)
  }
  + O(\h^2)
\end{split}
\label{eq:formal_fxht}
\end{equation}
Now we define $\partial_\h \mathbf x^\h_t|_{\h=0}$ as $\delta \mathbf x_t$ for simplicity, then by taking differential of \eqref{eq:formal_xht}, we get
\begin{equation}
\begin{split}
  \delta \mathbf x_t
  &=
  \partial_\h \mathbf x^\h_t|_{\h=0}
  \\
  &=
  \int_0^t
  \, 
  \left(
    \underbrace{
      \partial_\h
    \bm f^{\rm PF}_{\theta} (\mathbf x^\h_t,t)|_{\h=0}
    }_{
      (\partial_\h \mathbf x^\h_t|_{\h=0} \cdot \nabla) \bm f^{\rm PF}_{\theta} (\mathbf x_t,t)
    }
    -
    \frac{g(t)^2}{2}
    \left[
    \bm s_{\theta} ( \mathbf x_t, t) - \nabla \log q^{\h=0}_t(\mathbf x_t)
    \right]
  \right)
  \rd t
  \\
  &=
  \int_0^t
  \, 
  \left(
      (\delta \mathbf x_t \cdot \nabla) \bm f^{\rm PF}_{\theta} (\mathbf x_t,t)
    -
    \frac{g(t)^2}{2}
    \left[
    \bm s_{\theta} ( \mathbf x_t, t) - \nabla \log q^{\h=0}_t(\mathbf x_t)
    \right]
  \right)
  \rd t
  ,
\end{split}
\end{equation}
which is the integral of 
\begin{equation}
\begin{split}
  \dot{\delta \mathbf x}_t
  &=
  \underbrace{
  (\delta \mathbf x_t \cdot \nabla) \bm f^{\rm PF}_{\theta} (\mathbf x_t,t)
    -
    \frac{g(t)^2}{2}
    \left[
    \bm s_{\theta} ( \mathbf x_t, t) - \nabla \log q^{\h=0}_t(\mathbf x_t)
    \right]
  }_{
    \delta {\bm f}^{\rm PF}_{\theta}(\mathbf x_t, \delta \mathbf x_t, t)
  }
  .
\end{split}
\end{equation}
By combining it to $O(\h^0)$ ODE for $\mathbf x_t$, i.e., the probability flow ODE, we get \cref{eq:coupled-p-flow}.

Next, we apply the generic formula that is valid with arbitrary function $f(\mathbf x)$,
\begin{equation}
\begin{split}
  f(\mathbf x^\h_t)
  &=
  f(\mathbf x_t)
  +
  \h \delta \mathbf x_t \cdot \nabla f(\mathbf x_t)
  +
  O(\h^2)
  ,
\end{split}
\end{equation}
to the right hand side of the self-consistent equation of log-likelihood in \cref{thm:likelihood_wkb}.
\begin{equation}
\begin{split}
  &\log q^\h_0(\mathbf x_0)
  \\
  &=
  \log \pi(\mathbf x^\h_T)
  +
  \int_0^T 
  \, 
  \nabla \cdot
  \left(
  \bm f^{\rm PF}_{\theta} (\mathbf x^\h_t, t)
  - 
  \frac{\h g(t)^2}{2}
  [
  \bm s_{\theta} (\mathbf x^\h_t, t) - \nabla \log q^\h_t(\mathbf x^\h_t)
  ]
  \right)
  \rd t 
  \\
  &=
  \begin{array}{c}
  \log \pi(\mathbf x_T)
  \\
  +
  \\
  \h \delta \mathbf x_t \cdot \nabla \log \pi(\mathbf x_T)
  \end{array}
  +
  \int_0^T 
  \, 
  \left(
    \begin{array}{c}
    \nabla \cdot \bm f^{\rm PF}_{\theta} (\mathbf x_t, t)
    \\
    +
    \\
    \h (\delta \mathbf x_t \cdot \nabla)\nabla \cdot \bm f^{\rm PF}_{\theta} (\mathbf x_t, t)
    \end{array}
  - 
  \frac{\h g(t)^2}{2}
  \nabla \cdot
  [
  \bm s_{\theta} (\mathbf x_t, t) - \nabla \log q^{\h=0}_t(\mathbf x_t)
  ]
  )
  \right)
  +O(\h^2)
  ,
\end{split}
\end{equation}
then, this expression is equivalent to what we wanted to prove, i.e., \cref{eq:q-h-1}.

\section{ Computation of determinant } \label{sec:jacobian}

Let us detail on the computation of the determinant of an operator of the form,
\begin{equation}
\det \left( \p_t - M \right) .
\end{equation}
We first factorize this as 
$\det \left( \p_t - M \right)
= 
\det \p_t \cdot \det \left(1  - \p^{-1}_t M \right)
$. 
The latter factor is computed as 
\begin{equation}
\det \left(1  - \p^{-1}_t M \right)
= \exp \log  \det \left(1 - \p^{-1}_t M \right)
= \exp {\rm tr\,} \ln \left( 1  - \p^{-1}_t M \right)
= \exp {\rm tr\,}
\sum_{n=1}^\infty  
\left( - \frac{1}n (\p^{-1}_t M)^n \right) . 
\end{equation}
Noting that $\p^{-1}_t \delta (t-t') = \theta (t-t')$, 
the term with $n=1$ gives
\begin{equation}
\exp\left( - \theta (0) \int \rd t \,  {\rm tr\, } M \right),
\end{equation}
where $\theta(x)$ is the step function. 
Its value at $0$ depends on the discretization scheme as
\begin{equation}
\theta (0)  = 
\begin{cases}
0 & \text{Ito} \\
\frac{1}2 & \text{Stratonovich} \\
1 & \text{Reverse-Ito}
\end{cases} .
\label{eq:theta-0}
\end{equation}
All the higher-order terms vanish. For example, the term with $n=2$ reads 
\begin{equation}
 - \int 
 \theta (t-t') \theta (t' - t ) {\rm tr\,} M^2
\,  \rd t \rd t' 
 = 0.
\end{equation}
Thus, we have
\begin{equation}
\det \left( \p_t - M \right) 
\propto 
\exp\left( - \theta (0) \int \rd t \,  {\rm tr\, } M \right),
\end{equation}
with $\theta (0)$ given by \cref{eq:theta-0}.

\section{ Detail of the example in \cref{sec:1d-gauss} }\label{app:ex-gaussian}

We here describe the details of the simple example discussed in \cref{sec:1d-gauss}.

The data distribution is taken to be a one-dimensional Gaussian distribution,
\begin{equation}
P_0 (x_0) 
= 
\mathcal N (x_0 \,|\, 0, v_0) . 
\end{equation}
For the forward process, 
we use the following SDE,
\begin{equation}
 \rd x_t = - \frac{1}2 \beta x_t \, \rd t + \sqrt{\beta} \rd w_t. 
\end{equation}
Namely, we have chosen the force and noise strength as 
\begin{equation}
f(x) = - \frac{\beta}{2} x, 
\quad 
g = \sqrt{\beta} . 
\end{equation}
We here take $\beta$ to be constant.
In the current situation, the distribution stays Gaussian with a zero mean throughout the time evolution, and the distribution $p_t (x)$ is fully specified by its variance. 
Using Ito's formula, 
\begin{equation}
\rd (x_t^2) = 
2 x_t \, \rd x_t + \frac{1}2 \cdot 2 (\rd x_t)^2
= 
- \beta x_t^2 \, \rd t 
+ \beta \rd t . 
\end{equation}
Thus, the time evolution of the variance $v_t$ is described by 
\begin{equation}
\rd v_t =- \beta ( v_t - 1 )\rd t ,
\end{equation}
which can be solved 
with initial condition $v_{t=0}=v_0$ 
as 
\begin{equation}
v_t = 1 + e^{- \beta t} (v_0 - 1) . 
\label{eq:ex-var}
\end{equation}
The distribution at $t$ is given by 
\begin{equation}
p_t (x)  = \mathcal N (x \,|\, 0, v_t)  . 
\end{equation}
with $v_t$ given by Eq.~\eqref{eq:ex-var}.

\subsection{ Likelihood evaluation }

Suppose that the estimated score $s(x,t)$ is given by 
\begin{equation}
s(x,t)
=
\p_x \log q_t (x) = 
- \frac{x} {v_t} (1 + \epsilon) .
\end{equation}
where $\epsilon \in \mathbb R$ is a constant.
The parameters $\epsilon$ quantifies the deviation from the ideal estimation and 
when $\epsilon = 0$ we can recover the original data distribution perfectly. 
If $\epsilon \neq 0$, the likelihood $q_0 (x_0)$ depends nontrivially on parameter $\h$.
The force of the reverse process is written as
\begin{equation}
\wt{f}_{\h,\epsilon} (x)
=
f (x) - 
\frac{1+\h}{2} g^2 s (x,t)
=
- \frac{\beta}{2} x
- 
\frac{1+\h}{2} \beta 
\left(
- \frac{x} {v_t} (1 + \epsilon_t)
\right) 
=
\frac{\beta}2 
\left(
\frac{(1+\h)(1+\epsilon_t)}{v_t} - 1
\right)
x .
\end{equation}
Let us denote the variance of the model distribution $q_t^\h (x)$ by $v'_t$, which differs from $v_t$ when $\epsilon \neq 0$.
The variance $v'_t$ obeys 
\begin{equation}
\rd v'_t
= 
\beta 
\left(
\frac{(1+\h)(1+\epsilon)}{v_t} - 1
\right)
v'_t \rd t 
- 
\h \beta \rd t .
\end{equation}
If we solve this with the boundary condition $v'_T=v_T$, we have 
\begin{equation}
v'_t
= 
\frac{1}{(1+\h)(1+\epsilon)-1}
\left[
\h v_t + \epsilon(1+\h) 
e^{\beta (T-t)}
\left(
e^{- \beta (T-t) }
\frac{v_t}{v_T}
\right)^{(1+\h)(1+\epsilon)}
\right] .
\end{equation}
Note that, when $\epsilon=0$, we have $v'_t = v_t$ and it is independent of $\h$.
For nonzero $\epsilon$, $v'_0$ depends 
on $\h$ nontrivially.
The model distribution at $t$ is given by 
\begin{equation}
q^\h_t (x) = \mathcal N (x \, | \, 0, v'_t) . 
\end{equation}

%
The NLL can be expressed as 
\begin{equation}
- \mathbb E_{p_0(x_0)} [ \log q^\h_0 (x_0)]
= 
- 
\int \rd x 
\, 
\mathcal N ( x \,|\, 0 , v_0)
\log 
\mathcal N ( x \,|\, 0 , v'_{t=0})
=
\frac{1}2 
\left[
\log 
2\pi v'_{t=0} + \frac{v_0}{v'_{t=0}} 
\right] . 
\end{equation}

As another measure, we can compute the 
2-Wasserstein distance,
\begin{equation}
 W_2(p_0, p_T)^2 
= (\sqrt{v_0} - \sqrt{v'_{t=0}})^2 .
\end{equation}

Let us check the formula for the likelihood is satisfied. 
We shall check that the following formula is satisfied:
\begin{equation}
\log q^\h_0(x_0) 
- 
\log q^\h_T (x_T (x_0))
= 
\int_0^T \p_x 
\left[
\wt{f}^{\rm PF}  (x_t)
-
\frac{\h}2 g^2 
(s (x_t, t) - \nabla \ln q^\h_t (x_t))
\right]
\rd t ,
\end{equation}
where $\wt{f}^{\rm PF}(x) = \wt{f}_{\h=0,\epsilon}(x)$.
Noting that 
\begin{equation}
x_t = 
\sqrt{\frac{v'_t}{v_0}} x_0,
\end{equation}
and the LHS can be computed to give 
\begin{equation}
\text{LHS}    
= 
\frac{1}{2} \log \frac{v_T}{ v'_0 }.
\end{equation}
The integrand on the RHS reads 
\begin{equation}
\begin{split}
\p_x 
\left[ 
\wt{f}^{\rm PF}  (x)
-
\frac{\h}2 g^2 
(s (x,t) - \nabla \ln q^\h_t (x))
\right]
&=
\p_x 
\left[
- \frac{\beta}2 x 
- \frac{1+\h}2 \beta 
\left(
- \frac{x}{v_t} (1+\epsilon)
\right)
+
\frac{\h}{2} \beta 
\left(
- \frac{x}{v'_t}
\right) 
\right]
\\
&= 
\frac{\beta}{2} 
\left[
\frac{(1+\h)(1+\epsilon)}{v_t}
-1 
- \frac{\h}{v'_t} 
\right]. 
\end{split}
\end{equation}
The integration can be performed analytically, 
and we can check that the LHS indeed coincides with the RHS. 

\section{Setting of the pretraining in 2d synthetic data}
\label{app:2d_pretraining}

\subsection{Data}
\label{app:data}

\begin{figure}[t]
  \begin{tabular}{cc}
    \begin{minipage}[t]{0.45\hsize}
      \centering
      \centerline{\includegraphics[width=150px]{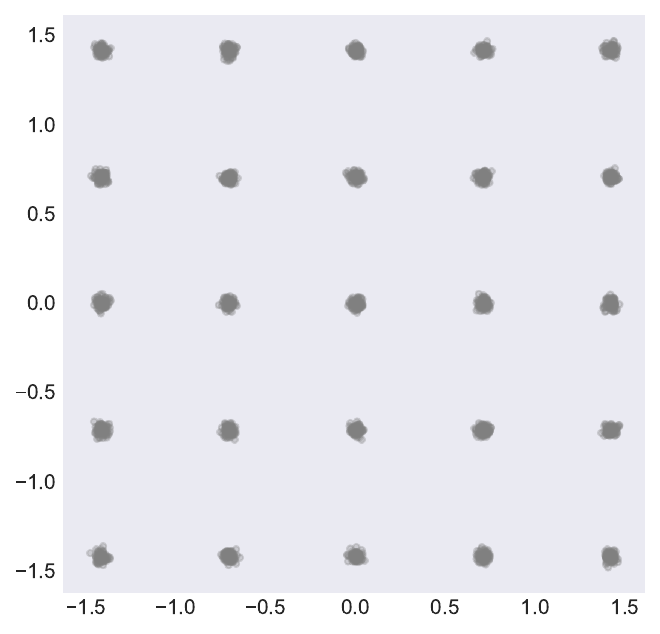}}
    \end{minipage} &
    \begin{minipage}[t]{0.45\hsize}
      \centering
      \centerline{\includegraphics[width=150px]{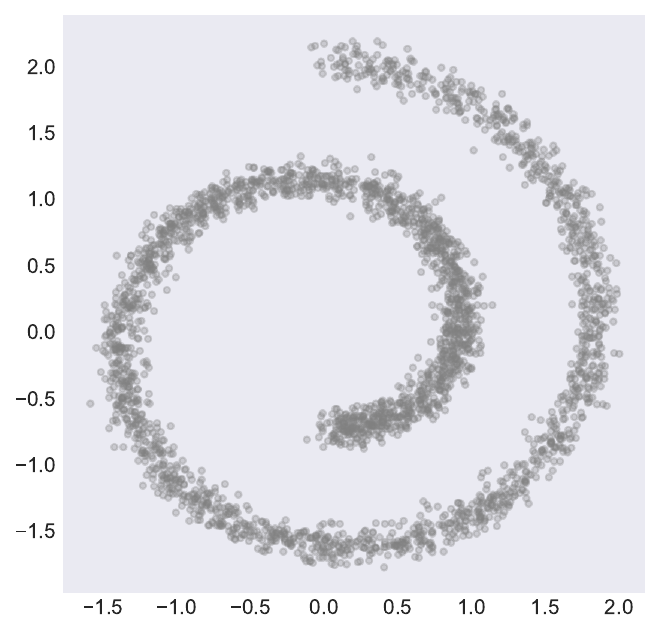}}
    \end{minipage}
  \end{tabular}
  \caption{\textbf{(Left)} 25-Gaussian data (3,000 samples), and \textbf{(Right)} Swiss-roll data (3,000 samples).}
  \label{fig:data}
\end{figure}

We use synthetic data similar to the synthetic data shown in \cref{fig:data} used in \cite{petzka2017regularization} to train our model.

Swiss-roll data is generated by {\tt sklearn.datasets.make\verb|_|swiss\verb|_|roll} \cite{scikit-learn} with {\tt noise = 0.5, hole = False}. 
This data itself is 3-dimensional data, so we project them to 2-dimensional by using {\tt [0, 2]} axes.
After getting data, we normalize it by its {\tt std} ($\approx$ {\tt 6.865}) to get data with {\tt std = 1}.

25-Gaussian data is generated by mixture of gaussians.
We first generate gaussian distributions with mean in $\{-4, 2, 0, 2, 4\}^2$, std = 0.05, and again divide each sample vector component by its {\tt std} ($\approx$ {\tt 2.828}) to get data with {\tt std = 1}.

\subsection{Training}
\label{app:training}

We used JAX \cite{jax2018github} and Flax \cite{flax2020github} to implement our score-based models with neural networks, and Optax \cite{deepmind2020jax} for the training.

As we write in the main part of this paper, we use simple neural network: $\mathbf (\mathbf x, t) \to {\tt concat}([\mathbf x, t]) \to \big({\tt Dense}(128) \to {\tt swish} \big)^3 \to {\tt Dense}(2) \to {\bm s}_\theta(\mathbf x, t)$ with default initialization both in training with Swiss-roll data and 25-Gaussian data.
The loss function 
is calculated by Monte-Carlo sampling in each training step:
\begin{enumerate}
  \item First we divide the time interval $[T_\text{min}, T_\text{max}]$ into 1,000 equal parts in advance to get a discretized diffusion time array.
  \item We take $N_\text{batch} =$ 512 batches of data $\{\mathbf x_i\}_{i=1,2,\dots,N_\text{batch}}$ and for each data point we take a discretized diffusion time $t_i$ uniformly from the array that we discretized in the first step.
  \item We compute the signal $\alpha(t_i)$ and noise $\sigma(t_i)$ at time $t_i$, computed from the definition of SDE, and take the following quantities as approximations of the loss function
  \begin{equation}
    \frac{1}{N_\text{batch}} \sum_{i=1}^{N_\text{batch}}
  \frac{g(t_i)^2}{2} 
  \left\|
  \frac{\mathbf x_{i, t} - \alpha(t_i)\mathbf x_i}{\sigma(t_i)^2}
  + {\bm s}_\theta(\mathbf x_{i, t}, t_i)
  \right\|^2 
  ,
  \label{eq:loss_MC}
  \end{equation}
  where $\mathbf x_{i, t} \sim \mathcal{N}(\alpha(t_i)\mathbf x_i, \sigma(t_i)^2 I)$ is the Monte-Carlo sample.
\end{enumerate}
The signal and noise functions $\alpha(t)$ and $\sigma(t)$ are:
\begin{equation}
\begin{split}
  &\text{\begin{sc}simple\end{sc}: }
  \quad
  \alpha(t) = e^{- \frac{\beta}{4}t^2}
  ,
  \quad
  \sigma(t)^2 = {1 - e^{- \frac{\beta}{2}t^2}}
  ,
  \\
  &\text{\begin{sc}cosine\end{sc}: }
  \quad
  \alpha(t) = \cos \left(\frac{\pi}{2} t\right)
  ,
  \quad
  \sigma(t)^2 = {1 - \cos \left(\frac{\pi}{2} t\right)}
  .
\end{split}
\end{equation}

These expressions are derived from general argument of SDE.
In general, once the SDE
\begin{equation}
  d \mathbf x_t 
  =
  f(t) \mathbf x + g(t) d \mathbf w_t
\end{equation}
is given, the conditional probability from time $t=0$ to $t$ is given by 
\begin{equation}
  p_{t|0}(\mathbf x_t|\mathbf x_0)
  =
  \mathcal{N}(\mathbf x_t | \alpha(t) \mathbf x_0, \sigma(t)^2 I),
\end{equation}
where
\begin{equation}
  \alpha(t)
  =
  e^{\int_0^t f(\xi) \rd \xi}
  ,
  \quad
  \sigma(t)^2 = \alpha(t)^2 \int_0^t \frac{g(\xi)^2}{\alpha(\xi)^2} \rd \xi
  ,
  \label{eq:sde_integral}
\end{equation}
which is essentially equivalent to the formula in \cite{karras2022elucidating}.

In training, 3,000 data points are taken in advance, and stochastic gradients are computed based on the Monte-Carlo loss function \eqref{eq:loss_MC} with batch size 512 mini-batches.
The gradients are used for optimization of the neural network ${\bm s}_\theta(\mathbf x_t, t)$ with Adam-optimizer with learning rate {\tt 1e-3} and default values determined by Optax \cite{deepmind2020jax}.
We train our models 16,000 epochs.

\subsection{Inference}
\begin{figure}[t]
  \begin{tabular}{cc}
    \begin{minipage}[t]{0.48\hsize}
      \centering
      \centerline{\includegraphics[width=230px]{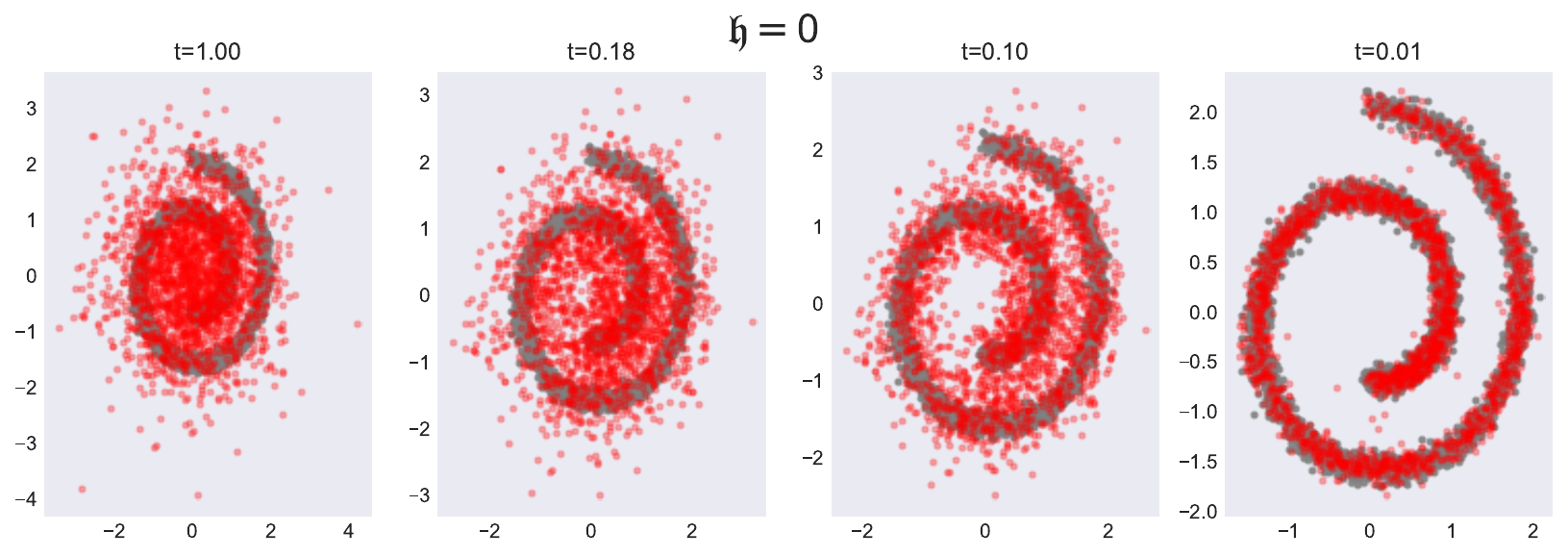}}
    \end{minipage} &
    \begin{minipage}[t]{0.45\hsize}
      \centering
      \centerline{\includegraphics[width=230px]{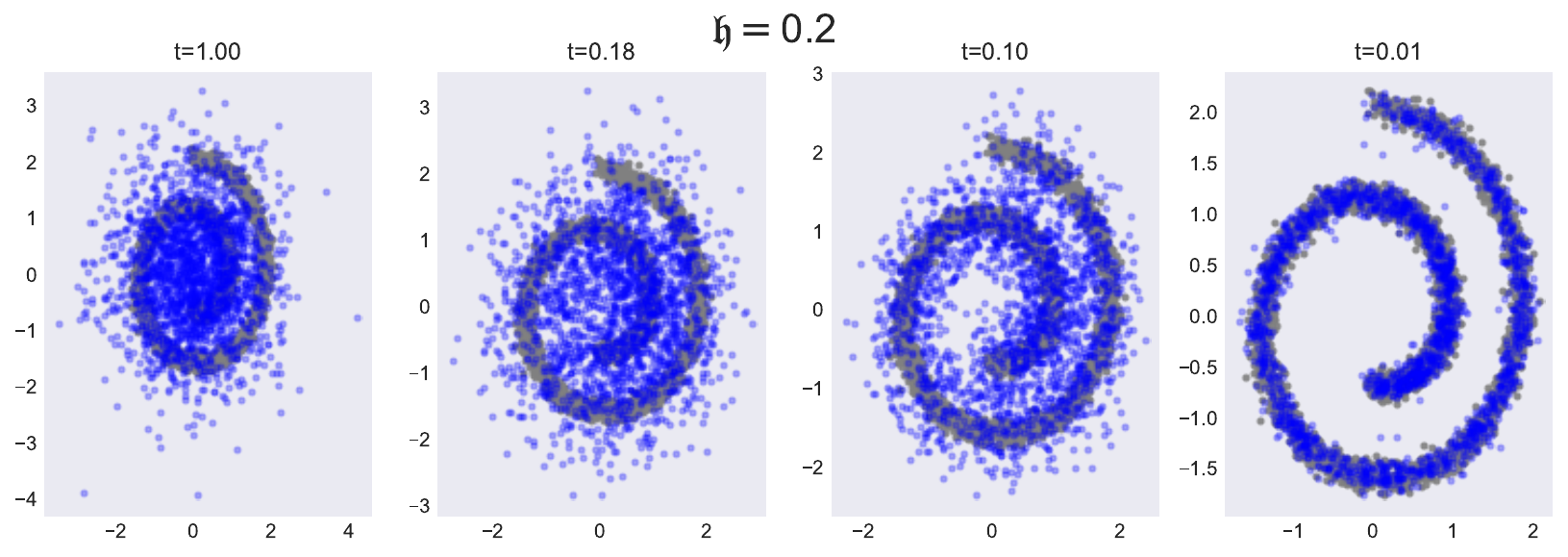}}
    \end{minipage}
  \end{tabular}
  \caption{\begin{sc}simple-sde\end{sc} pretrained model on Swiss-roll data with \textbf{(Left)} probability flow ODE ($\h=0$), \textbf{(Right)} SDE \eqref{eq:1para_gen_sde} with $\h=0.2$.}
  \label{fig:diff_sr_simple_gen}
  \begin{tabular}{cc}
    \begin{minipage}[t]{0.48\hsize}
      \centering
      \centerline{\includegraphics[width=230px]{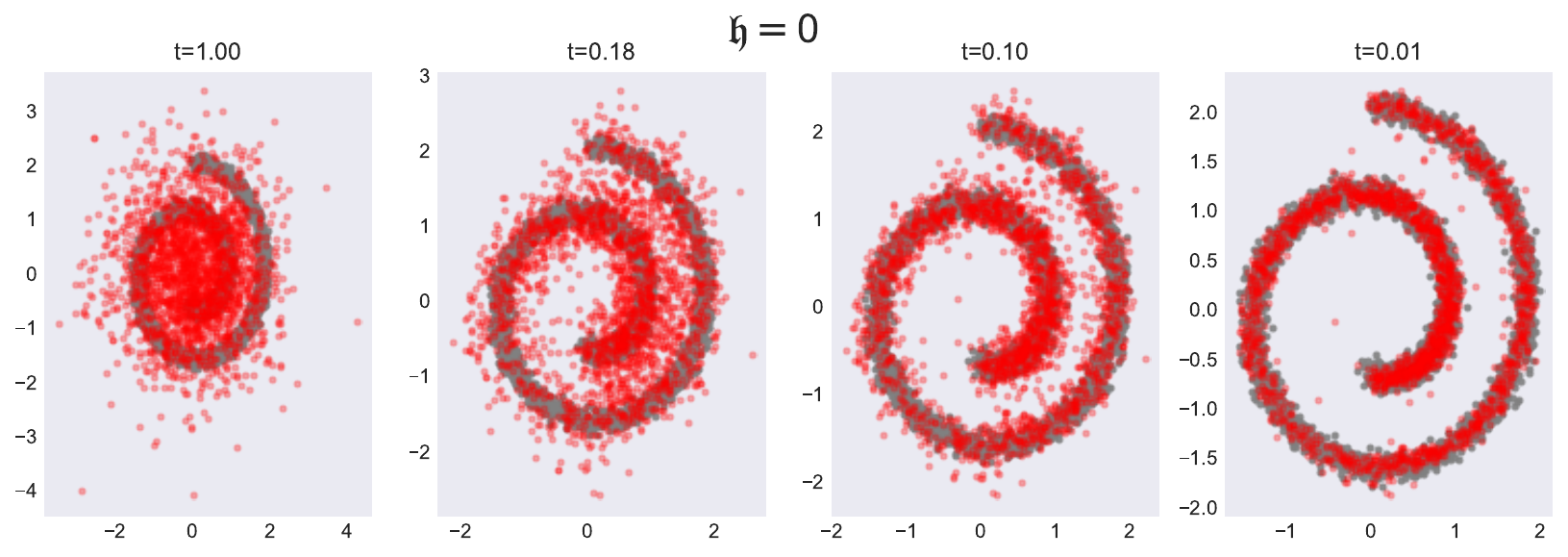}}
    \end{minipage} &
    \begin{minipage}[t]{0.45\hsize}
      \centering
      \centerline{\includegraphics[width=230px]{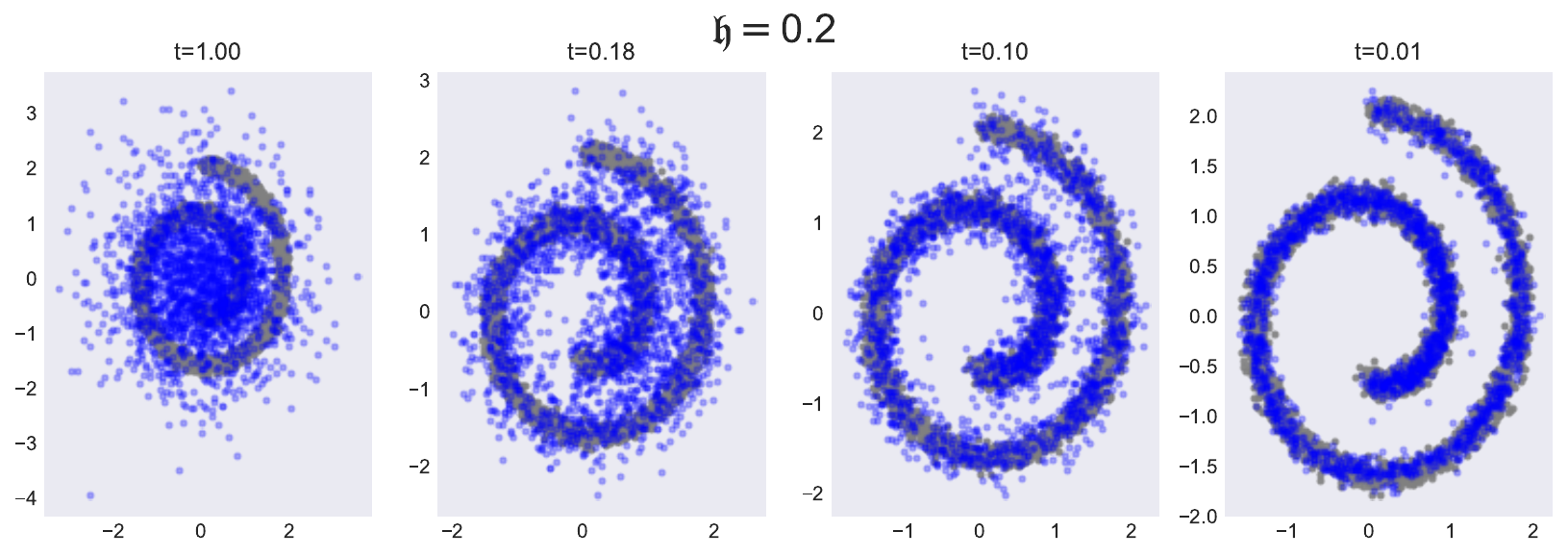}}
    \end{minipage}
  \end{tabular}
  \caption{\begin{sc}cosine-sde\end{sc} pretrained model on Swiss-roll data with \textbf{(Left)} probability flow ODE ($\h=0$), \textbf{(Right)} SDE \eqref{eq:1para_gen_sde} with $\h=0.2$.}
  \label{fig:diff_sr_cosine_gen}
  \begin{tabular}{cc}
    \begin{minipage}[t]{0.48\hsize}
      \centering
      \centerline{\includegraphics[width=230px]{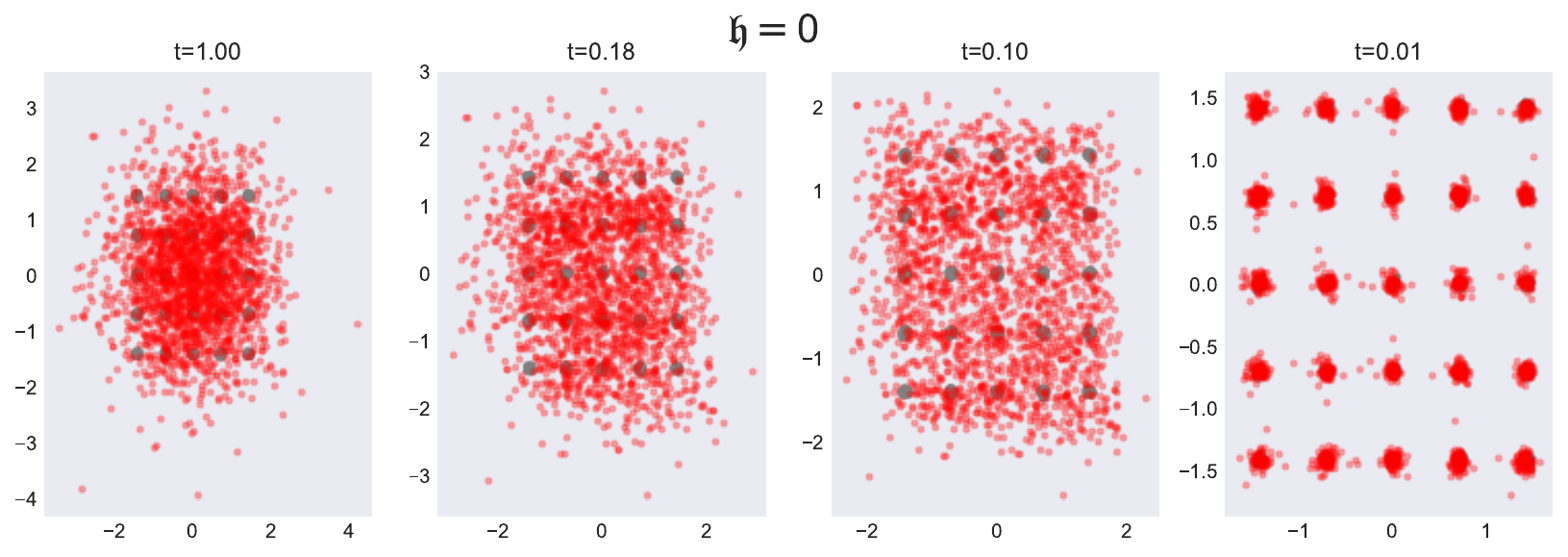}}
    \end{minipage} &
    \begin{minipage}[t]{0.45\hsize}
      \centering
      \centerline{\includegraphics[width=230px]{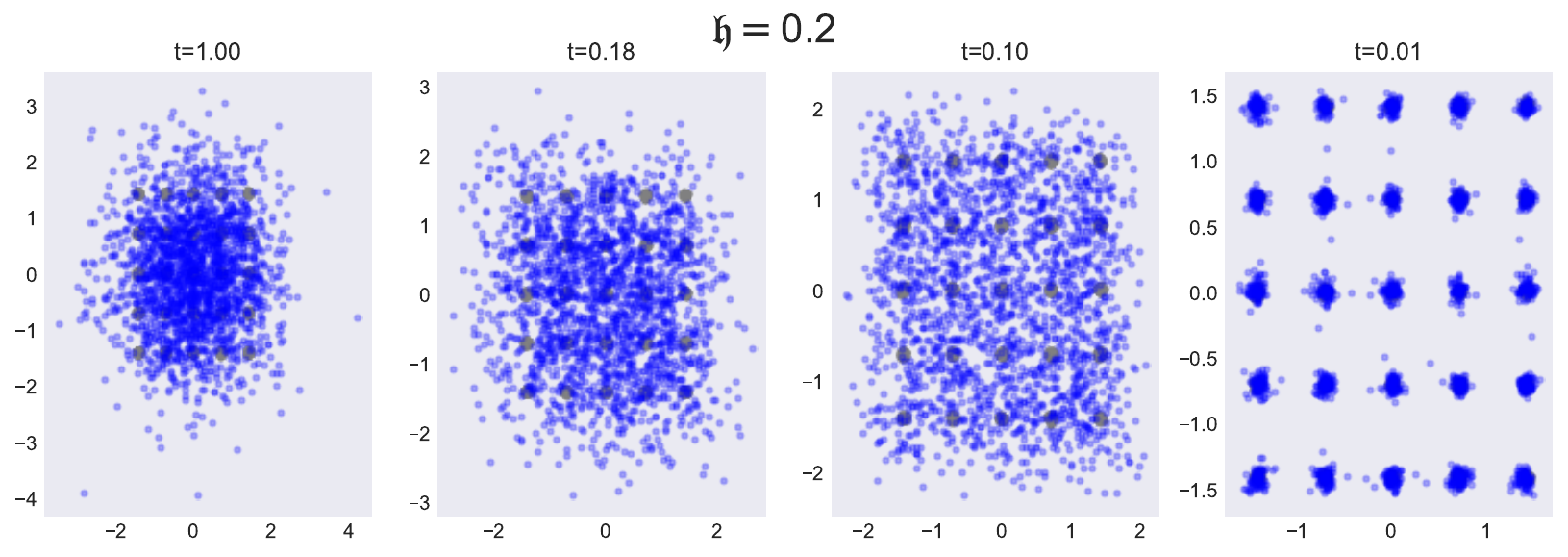}}
    \end{minipage}
  \end{tabular}
  \caption{\begin{sc}simple-sde\end{sc} pretrained model on 25-Gaussian data with \textbf{(Left)} probability flow ODE ($\h=0$), \textbf{(Right)} SDE \eqref{eq:1para_gen_sde} with $\h=0.2$.}
  \label{fig:diff_25_simple_gen}
\begin{tabular}{cc}
  \begin{minipage}[t]{0.48\hsize}
    \centering
    \centerline{\includegraphics[width=230px]{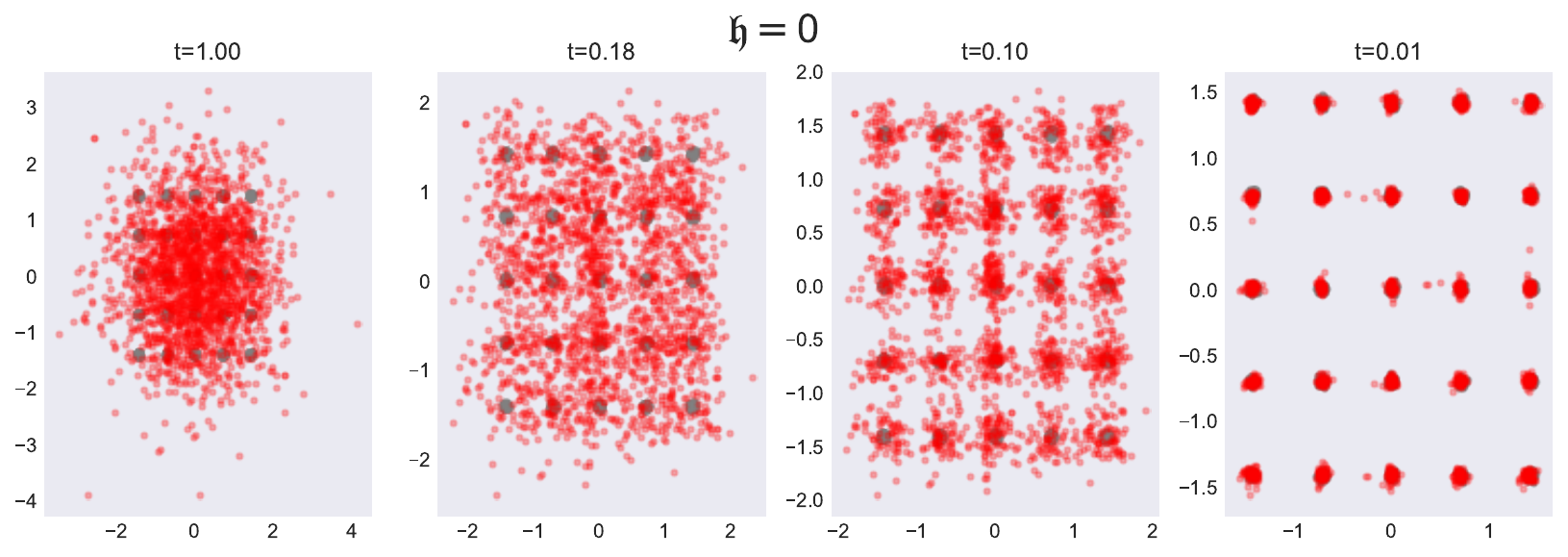}}
  \end{minipage} &
  \begin{minipage}[t]{0.45\hsize}
    \centering
    \centerline{\includegraphics[width=230px]{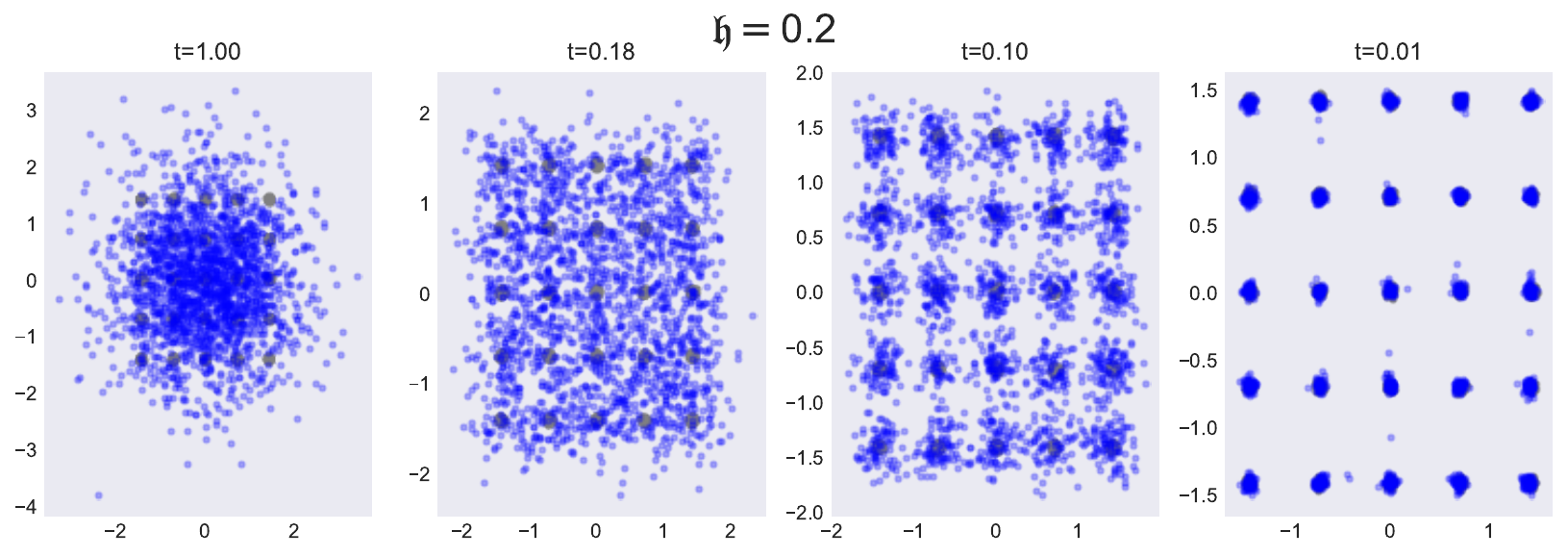}}
  \end{minipage}
\end{tabular}
\caption{\begin{sc}cosine-sde\end{sc} pretrained model on 25-Gaussian data with \textbf{(Left)} probability flow ODE ($\h=0$), \textbf{(Right)} SDE \eqref{eq:1para_gen_sde} with $\h=0.2$.}
\label{fig:diff_25_cosine_gen}
\end{figure}

Here we show some instances on generated data by our pretrained model in \cref{fig:diff_sr_simple_gen,fig:diff_sr_cosine_gen,fig:diff_25_simple_gen,fig:diff_25_cosine_gen}.
The figures are plotted with identical points in discretized time, $t=1,\ 0.18,\ 0.1,\ 0.01$.

\section{Derivative numerical error estimations in \cref{sec:2d-synthetic}}
\label{app:errors}

\subsection{Estimation of the local errors \eqref{eq:differential_errors}}
For simplicity, we omit ``$\h=0$'' script here.
In this notation, error of $\log q_t(\mathbf x)$ and local errors \eqref{eq:differential_errors} are
\begin{equation}
  \begin{split}
    \text{err} (N[\log q_t(\mathbf x)])
    &=
    \log q_t(\mathbf x)
    -
    N[\log q_t(\mathbf x)]
    ,
    \\
    \text{err} (\hat{\nabla} N[\log q_t(\mathbf x)])
    &=
    \nabla \log q_t(\mathbf x)
    -
    \hat{\nabla} N[\log q_t(\mathbf x)]
    ,
    \\
    \text{err} (\hat{\nabla}^2 N[\log q_t(\mathbf x)])
    &=
    \nabla^2 \log q_t(\mathbf x)
    -
    \hat{\nabla}^2 N[\log q_t(\mathbf x)]
    .
  \end{split}
\end{equation}
We use a third-party solver {\tt scipy.integrate.solve\verb|_|ivp} \cite{2020SciPy-NMeth}, and it has inputs {\tt atol} and {\tt rtol} that control the errors in the subroutine, to make it clear, let us call the numerical value as $N_{{\tt atol}, {\tt rtol}}[...]$.
It is plausible to expect the order of errors are same between numerical calculations based on the same order of tolerances, say 
\begin{equation}
\begin{split}
  O\Big(\text{err} (\hat{\nabla}^{0 \text{ or } 1\text{ or }2} N_{({\tt 1.1*atol}), ({\tt 1.1*rtol})}[\log q_t(\mathbf x)])\Big)
  =
  O\Big(
  \text{err} (\hat{\nabla}^{0 \text{ or } 1\text{ or }2} N_{{\tt atol}, {\tt rtol}}[\log q_t(\mathbf x)])
  \Big)
  ,
\end{split}
\end{equation}
where $O$ means order estimate.
Based on this observation, we can estimate each error.
For example,
\begin{equation}
\begin{split}
  &
  \underbrace{
    N_{({\tt 1.1*atol}), ({\tt 1.1*rtol})}[\log q_t(\mathbf x)]
  }_{
    \log q_t(\mathbf x) + \text{err} (N_{({\tt 1.1*atol}), ({\tt 1.1*rtol})}[\log q_t(\mathbf x)])
  }
  -
  \underbrace{
    N_{{\tt atol}, {\tt rtol}}[\log q_t(\mathbf x)]
  }_{
    \log q_t(\mathbf x) + \text{err} (N_{{\tt atol}, {\tt rtol}}[\log q_t(\mathbf x)])
  }
  \\
  &=
  \text{err} (N_{({\tt 1.1*atol}), ({\tt 1.1*rtol})}[\log q_t(\mathbf x)])
  -
  \text{err} (N_{{\tt atol}, {\tt rtol}}[\log q_t(\mathbf x)])
  \\
  &=
  O\Bigl(
    \text{err} (N_{{\tt atol}, {\tt rtol}}[\log q_t(\mathbf x)])
  \Bigr)
  .
  \label{eq:estimate_logq_error}
\end{split}
\end{equation}

\subsection{{\tt subtraction}: Estimation of the local errors \eqref{eq:differential_errors} by subtraction}
\label{app:sub}
On the order estimations for differential operators, we have 2 choices.
First choice is estimating them simply by 
\begin{equation}
    \text{err} (\hat{\nabla}^{1\text{ or }2} N[\log q_t(\mathbf x)])
  =
  \hat{\nabla}^{1\text{ or }2} N_{({\tt 1.1*atol}), ({\tt 1.1*rtol})}[\log q_t(\mathbf x)]
  -
  \hat{\nabla}^{1\text{ or }2}
  N_{{\tt atol}, {\tt rtol}}[\log q_t(\mathbf x)]
\end{equation}
We name error estimation scheme based on this by {\tt subtraction}.
this method is straightforward, however, we need doubled computation time, and we introduce more time-efficient estimation in the next section.

\subsection{{\tt model}: Estimation of the local errors \eqref{eq:differential_errors} by Taylor expansion and score models}
\label{app:model}
Next choice is simply based on the error
\begin{equation}
  N[\log q_t(\mathbf x)]
  =
  \log q_t(\mathbf x)
  +
  \text{err} (N[\log q_t(\mathbf x)])
  .
\end{equation}
By applying the discrete differential \eqref{eq:discrete_nabla}, we can get
\begin{equation}
\begin{split}
  \hat{\nabla} N[\log q_t(\mathbf x)]
  &=
  \left[
  \begin{array}{l}
    \frac{
      N[\log q_t(\mathbf x_t + [\Delta x, 0]^\top)]
      -
      N[\log q_t(\mathbf x_t - [\Delta x, 0]^\top)]
      }{2\Delta x}
    \\
    \frac{
      N[\log q_t(\mathbf x_t + [0, \Delta x]^\top)]
      -
      N[\log q_t(\mathbf x_t - [0, \Delta x]^\top)]
      }{2\Delta x}
  \end{array}
  \right]
  .
\end{split}
\end{equation}
It would be sufficient to show $x$ component:
\begin{equation}
\begin{split}
  &\hat{\partial}_x N [\log q_t(\mathbf x)]
  \\
  &=
  \frac{
      N[\log q_t(\mathbf x + [\Delta x, 0]^\top)]
      -
      N[\log q_t(\mathbf x - [\Delta x, 0]^\top)]
      }{2\Delta x}
  \\
  &=
  \frac{
      \log q_t(\mathbf x + [\Delta x, 0]^\top)
      -
      \log q_t(\mathbf x - [\Delta x, 0]^\top)
      }{2\Delta x}
  +
  \frac{
      \text{err} (N[\log q_t(\mathbf x + [\Delta x, 0]^\top)])
      - \text{err} (N[\log q_t(\mathbf x - [\Delta x, 0]^\top)])
      }{2\Delta x}
  \\
  &=
  \frac{
      \partial_x\log q_t(\mathbf x) 2 \Delta x 
      +
      O(\Delta x^3)
      }{2\Delta x}
      +
      \frac{O(\text{err} (N[\log q_t(\mathbf x)]))}{\Delta x}
      \\
  &=
  \partial_x\log q_t(\mathbf x)
  + O(\Delta x^2)
  + \frac{O(\text{err} (N[\log q_t(\mathbf x)]))}{\Delta x}
  ,
\end{split}
\end{equation}
where we use the facts: 1) $O(\Delta x^2)$ term vanishes from the numerator thanks to the symmetric descrete differential, 2) orders of $\text{err} (N[\log q_t(\mathbf x))]$ and $\text{err} (N[\log q_t(\mathbf x \pm [\Delta x, 0]^\top)])$ are same, in the 3rd line.
This fact is achieved by the Runge-Kutta integrator.
Therefore, by subtracting the true value from both side, we get 
\begin{equation}
  \text{err} (\hat{\nabla} N[\log q_t(\mathbf x)])
  =
  \nabla \log q_t(\mathbf x)
  -
  \hat{\nabla} N[\log q_t(\mathbf x)]
  =
  O(\Delta x^2) + \frac{O(\text{err} (N[\log q_t(\mathbf x)]))}{\Delta x}
\end{equation}
We can derive the error for $\nabla^2$ similarly:
\begin{equation}
  \small
\begin{split}
  &\hat{\nabla}^2 N[\log q_t(\mathbf x)]
  \\
  &=
  \frac{
    N[\log q_t(\mathbf x + [\Delta x, 0]^\top )]
     + N[\log q_t(\mathbf x - [\Delta x, 0]^\top)]
     + N[\log q_t(\mathbf x + [0, \Delta x]^\top)]
     + N[\log q_t(\mathbf x - [0, \Delta x]^\top)]
     -4N[\log q_t(\mathbf x)]
    }{\Delta x^2}
  \\
  &=
  \frac{
    \log q_t(\mathbf x + [\Delta x, 0]^\top )
     + \log q_t(\mathbf x - [\Delta x, 0]^\top)
     + \log q_t(\mathbf x + [0, \Delta x]^\top)
     + \log q_t(\mathbf x - [0, \Delta x]^\top)
     -4\log q_t(\mathbf x)
    }{\Delta x^2}
    \\&\quad
    +
    \frac{
      \begin{array}{r}
    \text{err} (N[\log q_t(\mathbf x + [\Delta x, 0]^\top )])
     + \text{err} (N[\log q_t(\mathbf x - [\Delta x, 0]^\top)])
     + \text{err} (N[\log q_t(\mathbf x + [0, \Delta x]^\top)])
     + \text{err} (N[\log q_t(\mathbf x - [0, \Delta x]^\top)])
     \\
     -4\text{err} (N[\log q_t(\mathbf x)])
      \end{array}
    }{\Delta x^2}
    \\
    &=
    \frac{
      \nabla^2 \log q_t(\mathbf x) \Delta x^2
      + O(\Delta x^4)
      }{\Delta x^2}
      +
      \frac{
      O(\text{err} (N[\log q_t(\mathbf x)]))
      }{\Delta x^2}
    \\
    &=
    \nabla^2 \log q_t(\mathbf x)
    + O(\Delta x^2)
      +
      \frac{
      O(\text{err} (N[\log q_t(\mathbf x)]))
      }{\Delta x^2}
      ,
\end{split}
\normalsize
\end{equation}
where we use that $O(\Delta x^3)$ term in the numerator vanishes because we use the 5-point approximation of the laplacian.
Now we get 
\begin{equation}
  \text{err} (\hat{\nabla}^2 N[\log q_t(\mathbf x)])
  =
  \nabla^2\log q_t(\mathbf x)
  -
  \hat{\nabla}^2 N[\log q_t(\mathbf x)]
  =
  O(\Delta x^2)
      +
      \frac{
      O(\text{err} (N[\log q_t(\mathbf x)]))
      }{\Delta x^2}
\end{equation}
In summary, we conclude each error as 
\begin{equation}
\begin{split}
  \text{err} (\hat{\nabla} N[\log q_t(\mathbf x)])
  &=
  {\bm C}^{(1)}_t(\mathbf x) \Delta x^2 + \frac{O(\text{err} (N[\log q_t(\mathbf x)])) \mathbf 1}{\Delta x}
  ,
  \\
  \text{err} (\hat{\nabla}^2 N[\log q_t(\mathbf x)])
  &=
  C^{(2)}_t (\mathbf x) \Delta x^2
      +
      \frac{O(\text{err} (N[\log q_t(\mathbf x)]))}{\Delta x^2}
  ,
\end{split}
\end{equation}
where $\mathbf 1$ is the vector with all 1 components, and ${\bm C}^{(1)}_t(\mathbf x), C^{(2)}_t(\mathbf x)$ are determined by the Taylor expansion remainder terms.
Typically, these are approximated by 
\begin{equation}
\begin{split}
  {\bm C}^{(1)}_t(\mathbf x)
  &\approx
  \text{3rd derivative of }\log q_t(\mathbf x)
  ,
  \\
  C^{(2)}_t(\mathbf x)
  &\approx
  \text{4th derivative of }\log q_t(\mathbf x)
  .
  \label{eq:reminders}
\end{split}
\end{equation}
Of course, we cannot access to the functions \eqref{eq:reminders}, however, it would be plausible to regard
\begin{equation}
  O(\nabla \log q_t(\mathbf x))
  =
  O(\bm s_{\theta} (\mathbf x, t))
  ,
\end{equation}
because the objective of the diffusion model training is to achieve $\bm s_{\theta} (\mathbf x, t) \approx \nabla \log p_t(\mathbf x)$ and in the ideal case, $q_t = p_t$.
By using this assumption, we regard 
\begin{equation}
\begin{split}
  {\bm C}^{(1)}_t(\mathbf x)
  &\approx
  -\nabla (\nabla \cdot \bm s_{\theta} (\mathbf x, t))
  \\
  C^{(2)}_t(\mathbf x)
  &\approx
  -\nabla^2 (\nabla \cdot \bm s_{\theta} (\mathbf x, t))
\end{split}
\label{eq:local_error_estimates}
\end{equation}
and estimate the local errors by 
\begin{equation}
\begin{split}
  \text{err} (\hat{\nabla} N[\log q_t(\mathbf x)])
  &\approx
  |\nabla (\nabla \cdot \bm s_{\theta} (\mathbf x, t)) \Delta x^2| + \left|\frac{O(\text{err} (N[\log q_t(\mathbf x)])) \mathbf 1}{\Delta x}\right|
  ,
  \\
  \text{err} (\hat{\nabla}^2 N[\log q_t(\mathbf x)])
  &\approx
  |\nabla^2 (\nabla \cdot \bm s_{\theta} (\mathbf x, t)) \Delta x^2|
      +
      \left|\frac{O(\text{err} (N[\log q_t(\mathbf x)]))}{\Delta x^2}\right|
  ,
  \label{eq:practical_errors}
\end{split}
\end{equation}
where $O(\text{err} (N[\log q_t(\mathbf x)]))$ can be estimated by using \eqref{eq:estimate_logq_error}.
We name error estimation scheme based on this by {\tt model}.

\subsection{Integral of the local errors}

Now we go back to \cref{thm:likelihood_perturbative} to explain our error estimation for $O(\h)$ correction to the (negative) log-likelihood. 
To calculate $O(\h)$ terms, we consider the paired ODE constructed by 
\begin{equation}
\begin{split}
  \dot{\mathbf x}_t &= \bm f^{\rm PF}_{\theta} (\mathbf x_t,t)
  ,
    \\
    \dot{\delta \mathbf x}_t & = \delta {\bm f}^{\rm PF}_{\theta}(\mathbf x_t, \delta \mathbf x_t, t)
    \quad
    (\text{by } \hat{\nabla} N[\log q_t(\mathbf x_t)])
    ,
    \\
    \dot{\delta \log q_t} & = \nabla \cdot \delta {\bm f}^{\rm PF}_{\theta}(\mathbf x_t, \delta \mathbf x_t, t)
    \quad
    (\text{by } \hat{\nabla}^2 N[\log q_t(\mathbf x_t)])
    ,
\end{split}
\end{equation}
as we wrote in \cref{alg:1st_logqsolver}.
In this expression, there is no numerical error in the RHS of $\dot{\mathbf x}_t$ except for the {\tt float32} precision that is the default of the deep learning framework.
In the RHS of $\dot{\delta \mathbf x}_t$, we have discretization of differential $\hat{\nabla}$ and numerical integral for $\log q^{\h=0}_t(\mathbf x_t)$, and as we noted, we omit $\h=0$ and call it as $\log q_t(\mathbf x_t)$.
By using the error estimation \eqref{eq:practical_errors}, we rewrite ODE for ${\delta \mathbf x}_t$ as
\begin{equation}
\begin{split}
  \dot{\delta \mathbf x}_t
  &=
  (\delta \mathbf x_t \cdot \nabla) \bm f^{\rm PF}_{\theta} (\mathbf x_t, t) - \frac{g(t)^2}{2} (
      \bm s_{\theta}(\mathbf x_t, t)
      - 
      \underbrace{
        \hat{\nabla} N\log q_t (\mathbf x_t) 
      }_{
        {\nabla} \log q_t (\mathbf x_t) 
        +
        \text{err} (\hat{\nabla} N[\log q_t(\mathbf x_t)])
      }
    )
  \\
  &=
  (\delta \mathbf x_t \cdot \nabla) \bm f^{\rm PF}_{\theta} (\mathbf x_t, t) - \frac{g(t)^2}{2} (
      \bm s_{\theta}(\mathbf x_t, t)
      - 
      {\nabla} \log q_t (\mathbf x_t))
  + \frac{g(t)^2}{2} 
     \text{err} (\hat{\nabla} N[\log q_t(\mathbf x_t)])
  .
\end{split}
\end{equation}
When we apply the solver, basically it discretize the system as
\begin{equation}
\begin{split}
  \delta \mathbf x_{t + \epsilon}
  =
  \delta \mathbf x_t
  + \epsilon \Big(
    (\delta \mathbf x_t \cdot \nabla) \bm f^{\rm PF}_{\theta} (\mathbf x_t, t) - \frac{g(t)^2}{2} (
      \bm s_{\theta}(\mathbf x_t, t)
      - 
      {\nabla} \log q_t (\mathbf x_t))
  + \frac{g(t)^2}{2} 
     \text{err} (\hat{\nabla} N[\log q_t(\mathbf x_t)])
  \Big)
  +
  \dots
  ,
\end{split}
\end{equation}
and truncate $\dots$ terms up to certain finite order, and apply this recursive equation from $\delta \mathbf x_0$.
For simplicity, we consider the liner term only here, and split $\delta \mathbf x_t = \delta \mathbf x_t^\text{true} + \text{err}^{(1)}_t$, then
\begin{equation}
  \begin{split}
    &
    \begin{pmatrix}
      \delta \mathbf x_{t + \epsilon}^\text{true}
      \\
      +
      \\
      \text{err}^{(1)}_{t + \epsilon}
    \end{pmatrix}
    \\
    &=
    \begin{pmatrix}
      \delta \mathbf x_{t}^\text{true}
      \\
      +
      \\
      \text{err}^{(1)}_{t}
    \end{pmatrix}
    + \epsilon \Big(
      \left(\begin{pmatrix}
        \delta \mathbf x_{t}^\text{true}
        \\
        +
        \\
        \text{err}^{(1)}_{t}
      \end{pmatrix} \cdot \nabla\right) \bm f^{\rm PF}_{\theta} (\mathbf x_t, t) - \frac{g(t)^2}{2} (
        \bm s_{\theta}(\mathbf x_t, t)
        - 
        {\nabla} \log q_t (\mathbf x_t))
    + \frac{g(t)^2}{2} 
       \text{err} (\hat{\nabla} N[\log q_t(\mathbf x_t)])
    \Big)
    \\
    &=
    \begin{pmatrix}
      \delta \mathbf x_t^\text{true}
    + \epsilon \Big(
      (\delta \mathbf x_t^\text{true} \cdot \nabla) \bm f^{\rm PF}_{\theta} (\mathbf x_t, t) - \frac{g(t)^2}{2} (
        \bm s_{\theta}(\mathbf x_t, t)
        - 
        {\nabla} \log q_t (\mathbf x_t))
      \Big)
      \\
      +
      \\
      \text{err}^{(1)}_{t}
    + \epsilon \Big(
      (\text{err}^{(1)}_{t} \cdot \nabla) \bm f^{\rm PF}_{\theta} (\mathbf x_t, t)
      +
      \frac{g(t)^2}{2} 
       \text{err} (\hat{\nabla} N[\log q_t(\mathbf x_t)])
       \Big)
    \end{pmatrix}
    ,
  \end{split}
\end{equation}
and we get the ODE for time $t$ error by taking continuous time limit:
\begin{equation}
  \dot{\text{err}}^{(1)}_{t}
  =
  (\text{err}^{(1)}_{t} \cdot \nabla) \bm f^{\rm PF}_{\theta} (\mathbf x_t, t)
      +
      \frac{g(t)^2}{2} 
       \text{err} (\hat{\nabla} N[\log q_t(\mathbf x_t)])
  .
\end{equation}
Note that $\text{err}^{(1)}_{t}$ is a vector with same dimension to $\mathbf x_t, \delta \mathbf x_t, \bm s_{\theta}$, and $\bm f^{\rm PF}_{\theta}$.
As same, we can get the ODE for $\dot{\delta \log q_t}$ term.
The discrete ODE is
\begin{equation}
\begin{split}
  &\delta \log q_{t+\epsilon}
  \\
  &=
  \delta \log q_t
  + \epsilon \Big(
    (\delta \mathbf x_t \cdot \nabla) \nabla \cdot \bm f^{\rm PF}_{\theta} (\mathbf x_t, t) - \frac{g(t)^2}{2} (
      \nabla \cdot \bm s_{\theta}(\mathbf x_t, t)
      - 
      {\nabla}^2 \log q_t (\mathbf x_t))
  + \frac{g(t)^2}{2} 
     \text{err} (\hat{\nabla}^2 N[\log q_t(\mathbf x_t)])
  \Big)
  +
  \dots
  ,
\end{split}
\end{equation}
and as same, by splitting true value and error value, we get
\begin{equation}
\begin{split}
  &
  \begin{pmatrix}
    \delta \log q_{t+\epsilon}^\text{true}
    \\
    +
    \\
    \text{err}^{(2)}_{t+\epsilon}
  \end{pmatrix}
  \\
  &=
  \begin{pmatrix}
    \delta \log q_{t}^\text{true}
    + \epsilon \Big(
      (\delta \mathbf x_t^\text{true} \cdot \nabla) \nabla \cdot \bm f^{\rm PF}_{\theta} (\mathbf x_t, t) - \frac{g(t)^2}{2} (
        \nabla \cdot \bm s_{\theta}(\mathbf x_t, t)
        - 
        {\nabla}^2 \log q_t (\mathbf x_t))
      \Big)
    \\
    +
    \\
    \text{err}^{(2)}_{t}
    + \epsilon \Bigl(
      (\text{err}^{(1)}_t \cdot \nabla) \nabla \cdot \bm f^{\rm PF}_{\theta} (\mathbf x_t, t)
      +
      \frac{g(t)^2}{2}
      \text{err} (\hat{\nabla}^2 N[\log q_t(\mathbf x_t)])
    \Bigr)
  \end{pmatrix}
  ,
\end{split}
\end{equation}
and get the ODE for $\text{err}^{(2)}_{t}$ as 
\begin{equation}
  \dot{\text{err}}^{(2)}_{t}
  =
  (\text{err}^{(1)}_t \cdot \nabla) \nabla \cdot \bm f^{\rm PF}_{\theta} (\mathbf x_t, t)
      +
      \frac{g(t)^2}{2}
      \text{err} (\hat{\nabla}^2 N[\log q_t(\mathbf x_t)])
  .
\end{equation}
In summary, we get the ODE system
\begin{equation}
\begin{split}
  \dot{\mathbf x}_t &= \bm f^{\rm PF}_{\theta} (\mathbf x_t,t)
  ,
    \\
    \dot{\delta \mathbf x}_t & = \delta {\bm f}^{\rm PF}_{\theta}(\mathbf x_t, \delta \mathbf x_t, t)
    \quad
    (\text{by } \hat{\nabla} N[\log q_t(\mathbf x_t)])
    ,
    \\
    \dot{\delta \log q_t} & = \nabla \cdot \delta {\bm f}^{\rm PF}_{\theta}(\mathbf x_t, \delta \mathbf x_t, t)
    \quad
    (\text{by } \hat{\nabla}^2 N[\log q_t(\mathbf x_t)])
    ,
    \\
    \dot{\text{err}}^{(1)}_{t}
  &=
  (\text{err}^{(1)}_{t} \cdot \nabla) \bm f^{\rm PF}_{\theta} (\mathbf x_t, t)
      +
      \frac{g(t)^2}{2} 
       \text{err} (\hat{\nabla} N[\log q_t(\mathbf x_t)])
    ,
    \\
    \dot{\text{err}}^{(2)}_{t}
  &=
  (\text{err}^{(1)}_t \cdot \nabla) \nabla \cdot \bm f^{\rm PF}_{\theta} (\mathbf x_t, t)
      +
      \frac{g(t)^2}{2}
      \text{err} (\hat{\nabla}^2 N[\log q_t(\mathbf x_t)])
  .
\end{split}
\end{equation}
The \begin{sc}errors\end{sc} in \cref{tab:nll} are calculated by solving more conservative (or over-estimated) ODE:
\begin{equation}
\begin{split}
  \dot{\text{err}}^{(1)}_{t}
  &=
  \Big|(\text{err}^{(1)}_{t} \cdot \nabla) \bm f^{\rm PF}_{\theta} (\mathbf x_t, t)\Big|
  +
  \Big|\frac{g(t)^2}{2} 
       \text{err} (\hat{\nabla} N[\log q_t(\mathbf x_t)])
  \Big|
    ,
    \\
    \dot{\text{err}}^{(2)}_{t}
  &=
  \Big|(\text{err}^{(1)}_t \cdot \nabla) \nabla \cdot \bm f^{\rm PF}_{\theta} (\mathbf x_t, t)
  \Big|
  +
  \Big|\frac{g(t)^2}{2}
      \text{err} (\hat{\nabla}^2 N[\log q_t(\mathbf x_t)])
  \Big|
  ,
\end{split}
\end{equation}
by using approximations \eqref{eq:practical_errors} and the error final value for $\log q_T$ as
\begin{equation}
  |\text{err}^{(1)}_T \cdot \nabla \log \pi(\mathbf x_T)|
  +
  |\text{err}^{(2)}_{T}|
  \label{eq:final_error}
\end{equation}

\subsection{Visualization of local error estimates }
\label{app:vis_errors}

\begin{figure}[t]
  \begin{center}
  \centerline{\includegraphics[width=0.9\columnwidth]{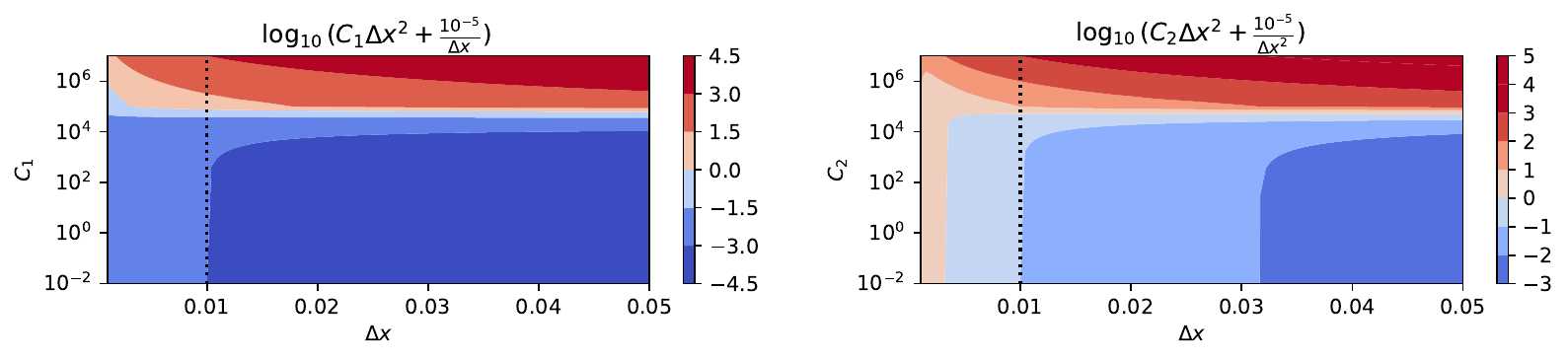}}
  \caption{ The order of local errors. \textbf{(Left)} Contour plot corresponding to $\text{err} (\hat{\nabla} N[\log q_t(\mathbf x_t)])$. \textbf{(Right)} Contour plot corresponding to $\text{err} (\hat{\nabla}^2 N[\log q_t(\mathbf x_t)])$, based on \eqref{eq:local_error_estimates}.
  Blue region has negative power, and relatively safe. Red region has positive power, and dangerous. The dotted line corresponds $\Delta x=0.01$ that is the value used in \cref{tab:nll}.
  The value $10^{-5}$ in the 2nd-term numerator is the typical tolerance value of {\tt 0th-logqSolver} in \cref{alg:1st_logqsolver}, that corresponds to $\text{err} (N[\log q_t(\mathbf x)])$.
  }
  \label{fig:errors}
  \centerline{\includegraphics[width=0.9\columnwidth]{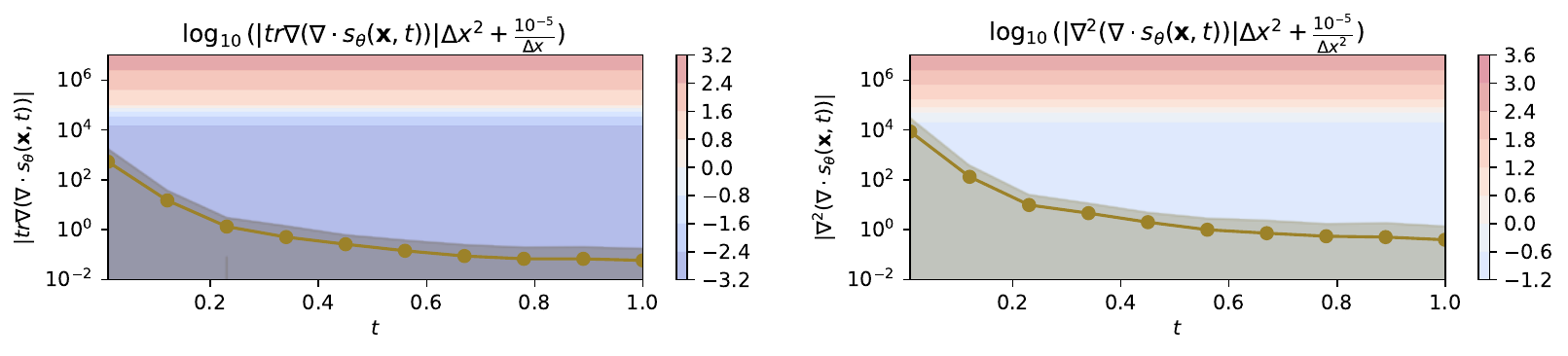}}
  \caption{Local error mean/std plot with pretrained model with \begin{sc}simple-SDE\end{sc} trained by Swiss-roll.}
  \label{fig:lerr_sr_simple}
  \centerline{\includegraphics[width=0.9\columnwidth]{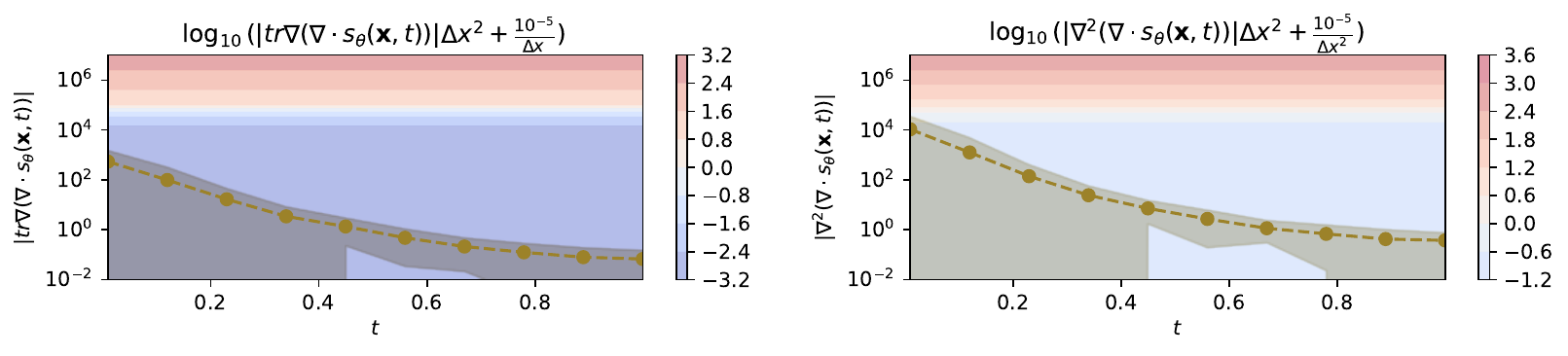}}
  \caption{Local error mean/std plot with pretrained model with \begin{sc}cosine-SDE\end{sc} trained by Swiss-roll.}
  \label{fig:lerr_sr_cosine}
  \centerline{\includegraphics[width=0.9\columnwidth]{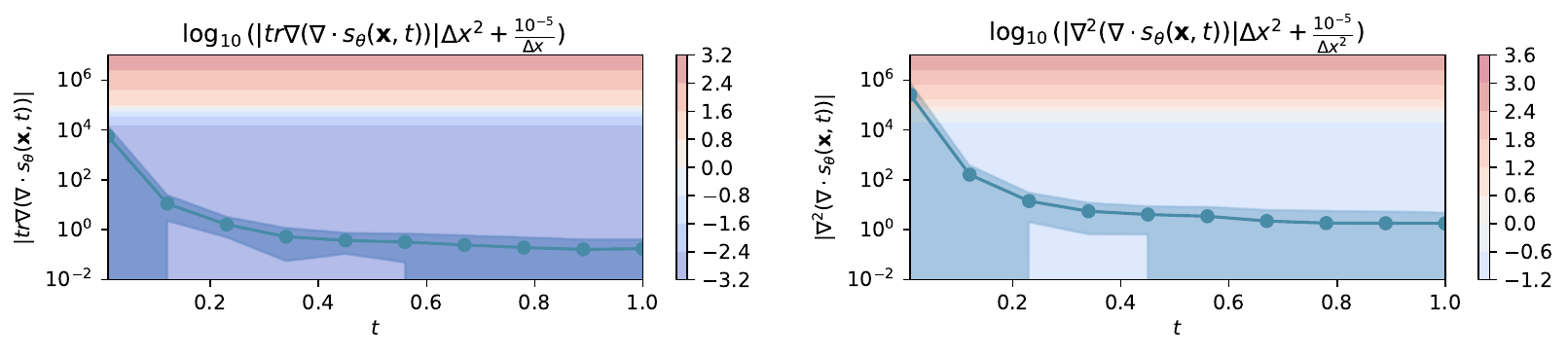}}
  \caption{Local error mean/std plot with pretrained model with \begin{sc}simple-SDE\end{sc} trained by 25-Gaussian.}
  \label{fig:lerr_25_simple}
  \centerline{\includegraphics[width=0.9\columnwidth]{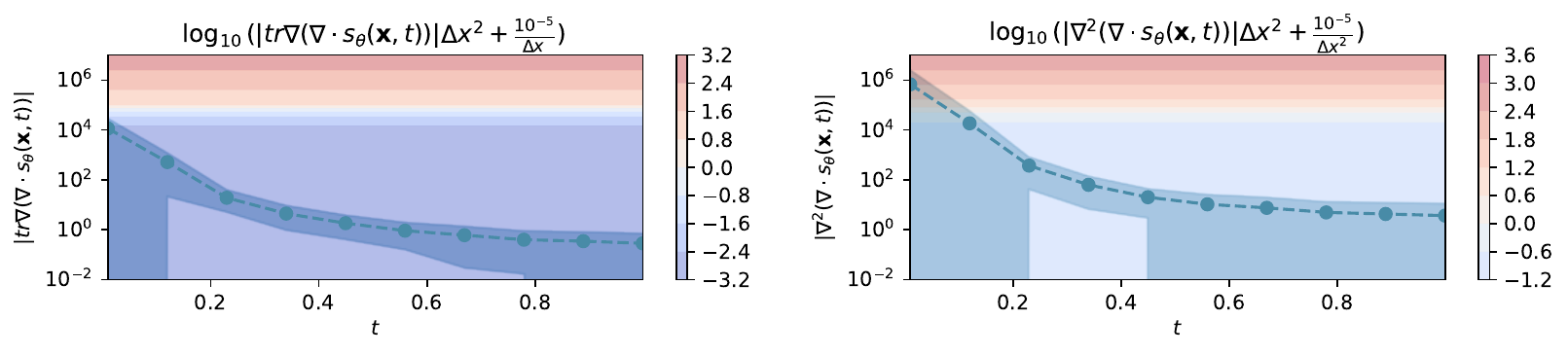}}
  \caption{Local error mean/std plot with pretrained model with \begin{sc}cosine-SDE\end{sc} trained by 25-Gaussian.}
  \label{fig:lerr_25_cosine}
  \end{center}
  \vskip -0.2in
\end{figure}

Of course, the degree of final error \eqref{eq:final_error} strongly depends on the order of the local errors $\text{err} (\hat{\nabla} N[\log q_t(\mathbf x_t)])$ and $\text{err} (\hat{\nabla}^2 N[\log q_t(\mathbf x_t)])$.
To see its order, we plot $\log_{10}$ scales of the local error functions in \cref{fig:errors}.

If we believe $\text{err} (N[\log q_t(\mathbf x)])$ is suppressed within the small tolerance value, we take it as $10^{-5}$ by the Runge-Kutta algorithm, the order of local errors are depending on the coefficients of $\Delta x^2$.
To check these values, we plot mean and std values of $\text{tr} \nabla (\nabla \cdot {\bm s}_\theta(\mathbf x, t))$ and $\nabla^2 (\nabla \cdot {\bm s}_\theta(\mathbf x, t))$ by sampling 500 points $\mathbf x \sim \text{Uniform}(\min({\tt validation\ set}) - 0.1, \max({\tt validation\ set}) - 0.1)$ at each time $t$ in \cref{fig:lerr_sr_simple,fig:lerr_sr_cosine,fig:lerr_25_simple,fig:lerr_25_cosine}.
Simultaneously, we plot colored contours that corresponds 
the value of local error estimates based on \eqref{eq:local_error_estimates} with $\Delta x=0.01$, that are exactly same as the values on dashed line in \cref{fig:errors}.

From these figures, one can see that the estimated local errors are almost located at safe region, i.e., errors are negative in log scale.
As one can see, the 25-Gaussian case \cref{fig:lerr_25_simple,fig:lerr_25_cosine}, the maximum values are slightly inside red regions, and we may be careful about it, but we leave further study as a future work.

\end{document}